\documentclass[10pt,twocolumn,letterpaper]{article}
\usepackage{makecell}
\usepackage{multirow}
\usepackage{arydshln}
\usepackage{amsthm}
\newtheorem{Proposition}{Observation}
\usepackage{cvpr}              % To produce the CAMERA-READY version
\definecolor{cvprblue}{rgb}{0.21,0.49,0.74}
\usepackage[pagebackref,breaklinks,colorlinks,allcolors=cvprblue]{hyperref}

\def\paperID{918} % *** Enter the Paper ID here
\def\confName{CVPR}
\def\confYear{2025}

\title{BHViT: Binarized Hybrid Vision Transformer}

\author{
Tian Gao$^{1,2}$\hspace{0.02in}
Zhiyuan Zhang$^{3}$ \hspace{0.02in}
Yu Zhang$^{4}$ \hspace{0.02in}
Huajun Liu$^{2}$ \hspace{0.02in}
Kaijie Yin$^{1}$ \hspace{0.02in}
Chengzhong Xu$^{1}$ \hspace{0.02in}
Hui Kong$^{1,}$ \thanks{Corresponding author}
\vspace{0.1in}\\
$^{1}$University of Macau \quad
$^{2}$Nanjing University of Science and Technology
\\
$^{3}$Singapore Management University
\quad
$^{4}$Shanghai Jiaotong University 
}

\begin{document}
\maketitle
\begin{abstract}
Model binarization has made significant progress in enabling real-time and energy-efficient computation for convolutional neural networks (CNN), offering a potential solution to the deployment challenges faced by Vision Transformers (ViTs) on edge devices. However, due to the structural differences between CNN and Transformer architectures, simply applying binary CNN strategies to the ViT models will lead to a significant performance drop. To tackle this challenge, we propose BHViT, a binarization-friendly hybrid ViT architecture and its full binarization model with the guidance of three important observations. Initially, BHViT utilizes the local information interaction and hierarchical feature aggregation technique from coarse to fine levels to address redundant computations stemming from excessive tokens. Then, a novel module based on shift operations is proposed to enhance the performance of the binary Multilayer Perceptron (MLP) module without significantly increasing computational overhead. In addition, an innovative attention matrix binarization method based on quantization decomposition is proposed to evaluate the token's importance in the binarized attention matrix. Finally, we propose a regularization loss to address the inadequate optimization caused by the incompatibility between the weight oscillation in the binary layers and the Adam Optimizer. Extensive experimental results demonstrate that our proposed algorithm achieves SOTA performance among binary ViT methods. The source code is released at: \href{https://github.com/IMRL/BHViT}{https://github.com/IMRL/BHViT}. 
\end{abstract}   
\vspace{-10pt}
\section{Introduction}
\label{intro}
\begin{figure}
\setlength{\belowcaptionskip}{-15pt}
\centering
\includegraphics[width=2.2in]{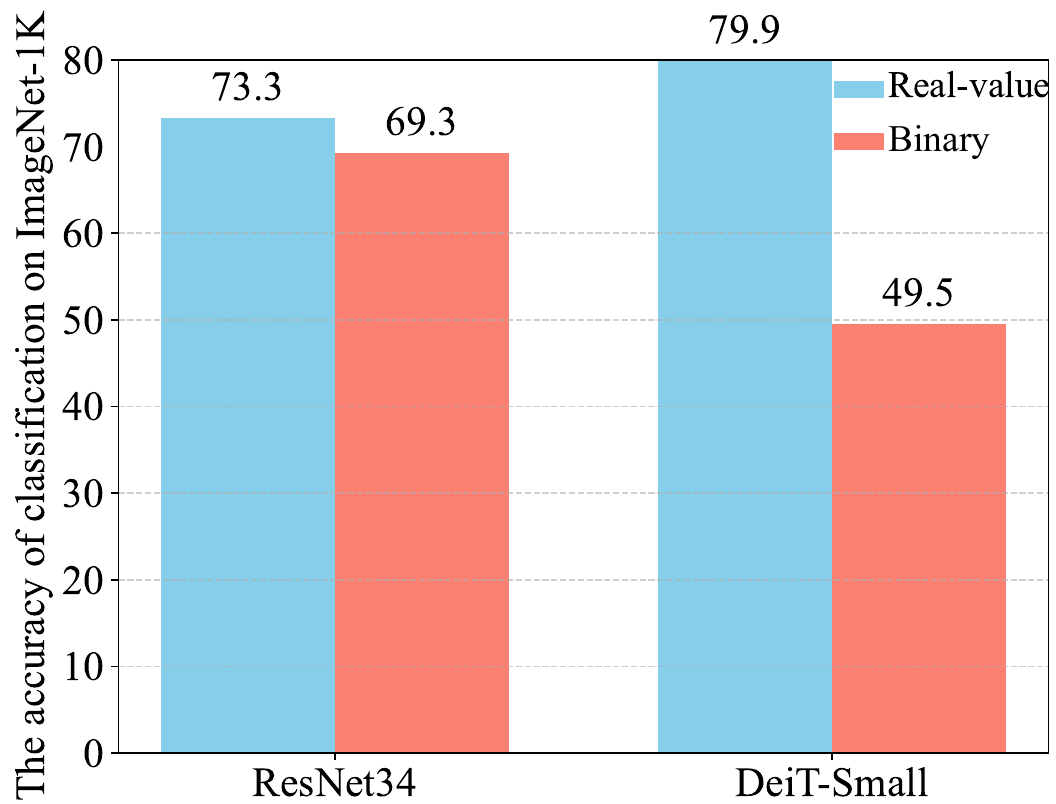}
\caption{The classification accuracy on ImageNet-1K dataset of the binary method, ReActNet~\cite{liu2020reactnet}, when applied in CNN and ViT architectures, respectively.}
\label{figduibik}
\end{figure}

In recent years, ViTs have made significant progress in many computer vision fields~\cite{touvron2021training,han2021transformer}. Nevertheless, due to the substantial model size and high computational complexity, deploying ViT in real-time application scenarios with limited computing resources is challenging. To deal with this issue, model quantization methods~\cite{li2022q,jung2021quantization} have been proposed, among which binary ViT techniques can be the most efficient ones. Especially, the very recent Large Language Models (LLMs)~\cite{achiam2023gpt} and Visual Language Models (VLM)~\cite{radford2021learning} all adopt the 
transformer architecture. Therefore, the exploration of binary ViT with high performance holds practical significance.
%The model binarization methods generally consist of binary CNN and ViT architectures. Despite the significant advancements in the binary CNN~\cite{chen2021bnn,liu2018bi,xu2021recu,rastegari2016xnor}, 
Although many binary CNN models have been preceding binary ViT, as shown in Fig.~\ref{figduibik}, directly applying existing binary CNN techniques (e.g., RSign and RPReLU~\cite{liu2020reactnet}) to the ViT framework can not solve the significant performance degradation. 
%Through the exploration in previous work~\cite{li2022q,he2022bivit,li2024bi,gao2024gsb}, it has been observed that two factors mostly affect the performance of binary ViT models.

The primary reasons for the performance degradation lie in two aspects. First, the back-propagation of the attention module can be easily corrupted by the multiple clip functions and the non-differentiability of the sign operator, resulting in vanishing gradients for most elements in the activation. Second, the binary attention matrix cannot accurately represent the differences among the similarities of different tokens. The binarized attention can introduce a lot of noise and reduce the signal-to-noise ratio of the attention matrix~\cite{ye2024differential}, leading to a performance drop.

To tackle these challenges, we propose a hybrid ViT framework in this paper, which is better suited for binarization than the ViT and its full binarization version. Our contributions can be summarized as follows,\\ 
\indent $\bullet$ We explore the reasons that cause the severe performance degradation of the current binarized ViT models. \\
\indent $\bullet$ Based on our research, we propose three novel modules to construct a high-performing binarization-friendly hybrid ViT framework. Meanwhile, we propose a regularization loss to address the incomplete optimization caused by the incompatibility between the weight oscillation and the Adam Optimizer.\\
\indent $\bullet$ To enhance the performance of the binary attention module, we propose a binarization scheme called Quantization Decomposition (QD) for attention matrices. \\
\indent $\bullet$ We have applied the BHViT to both classification and segmentation tasks and achieved SOTA performance.
\begin{figure}

\setlength{\belowcaptionskip}{-15pt}
\centering
\includegraphics[width=3.3in]{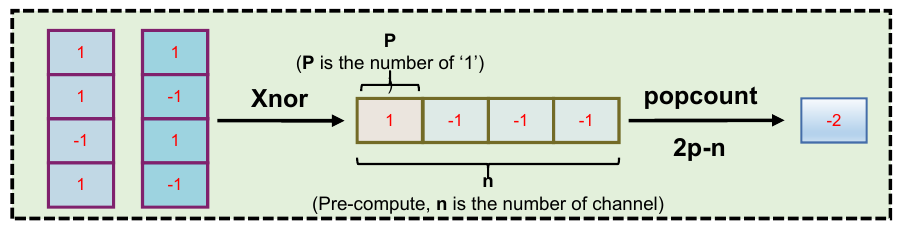}
\caption{The multiplication between binary vectors can be implemented by the Xnor and popcount. The result is $2p-n$}
\label{figxnor}
\end{figure}
\section{Related work}
\textbf{Binary Neural Networks (BNN).}~~Model binarization initially begins with the CNN framework. Courbariaux et al.~\cite{courbariaux2015binaryconnect} proposed the BinaryConnect model with binary weights and full-precision activations. To address the issue of non-differentiable sign operators, they introduced the $clip$ function to approximate the gradient during the back-propagation process. Based on  BinaryConnect, Hubara et al.~\cite{hubara2016binarized} further extended the binarization to the model's activation, allowing the BNN to utilize $Xnor$ and $popcount$ operations to approximate matrix multiplication. Rastegari et al.~\cite{rastegari2016xnor} introduced the scaling factor for both weight and activation to enhance BNN performance on large datasets~\cite{deng2009imagenet}. Lin et al.~\cite{lin2017towards} introduced multiple binarization bases for weights and activations to approximate the corresponding full-precision values, but it introduces additional computations. Liu et al.~\cite{liu2018bi} introduced residual links for each binary convolution layer to improve the representation capacity of the features. Furthermore, Liu et al.~\cite{liu2020reactnet} incorporated an $\mathrm{RPReLU}$ activation function and the $\mathrm{RSign}$ operator within a MobileNet-based~\cite{sandler2018mobilenetv2} architecture to reduce the performance gap between BNN and the corresponding full precision network. 

\noindent\textbf{Binary Vision Transformers (ViTs).} While binary operations (e.g., RSign~\cite{liu2020reactnet}) can be directly applied to ViTs, performance often degrades significantly. To address this, Li et al.~\cite{li2024bi} introduced the head-wise scaling operation to the attention module and utilized a ranking-aware distillation loss to train the binary ViT. Wang et al.~\cite{wang2024bvt} proposed channel-wise scale factors based on the Hamming distance between the binary $Q$ and $K$ tensors. Gao et al.~\cite{gao2024gsb} utilized distinct thresholds to transform the full-precision attention matrix and $V$ matrix into a superposition form consisting of multiple binary bases. Other approaches relax bit-level constraints for specific network components. He et al.~\cite{he2023bivit} proposed a mixed-precision vision transformer, binarizing the attention module and MLP weights while keeping MLP activations in full precision. Similarly, Lu et al.~\cite{lu2023understanding} maintained full-precision attention matrices in their mixed-precision ViT. 

In contrast to the above binary ViT methods, our work aims to mitigate the incompatibility between the ViT architecture and existing binarization techniques. Compared with BiReaL-Net~\cite{liu2018bi}, we first elucidate the significance of dense residual connections for binarizing the ViT architecture from an optimization perspective. Additionally, we slightly modify the original dense residual connections to accommodate the unique structural modules in Binary ViT.
\section{Background}
In the binary model, the weights and activations are constrained to two states, 1 and -1 (or 0 and 1). The multiplication of binary vectors is efficiently implemented using $Xnor$ and $popcount$ operations (Fig.~\ref{figxnor}).
The linear layer is a fundamental component of the ViT that accomplishes the channel-wise aggregation of feature information. Its binarization process is described by:
\vspace{-5pt}
\begin{equation}
\begin{aligned}
\label{blinear}
\mathbf{Y} =\mathrm{B}_\mathrm{Lin}\left( \mathbf{X},\mathbf{W} \right)= \alpha _{\mathbf{W}} \left( \hat{\mathbf{W}} \circledast \hat{\mathbf{X}} \right) ,
\end{aligned}
\end{equation}
where $\mathbf{Y}$, $\mathbf{X}$, $\mathbf{W}$, $\hat{\mathbf{W}}$, and $\hat{\mathbf{X}}$ represent the output, input, weight, binarized weight, and binarized activation of the linear layer, respectively. $\alpha _{\mathbf{W}}$ is a scaling factor. The symbol $\circledast$ denotes binary matrix multiplication implemented by $Xnor$ and $popcount$ operations.
To obtain $\hat{\mathbf{X}}$, as shown in Eq.~\ref{eq2-1}, the $sign$ operator and the piecewise polynomial function~\cite{liu2018bi} are applied to the forward and backward processes of the activation binarization, respectively.
\begin{equation}
\label{eq2-1}
\begin{aligned}
&\mathbf{Forward}~\hat{\mathbf{X}}=\mathrm{B}_a\left( \mathbf{X},a,b \right) =sign\left( \frac{\mathbf{X}-b}{a} \right) ,\\
	&\mathbf{Backward}~\frac{\partial L}{\partial \mathbf{X}}=\frac{\partial L}{\partial \hat{\mathbf{X}}}\frac{\partial \hat{\mathbf{X}}}{\partial \mathbf{X}}=\\
 &\left\{ \begin{matrix}
	\frac{\partial L}{\partial \hat{\mathbf{X}}}.\left( 2+2\left( \frac{\mathbf{X}-b}{a} \right) \right)&		b-a \leqslant \mathbf{X}<b\\
	\frac{\partial L}{\partial \hat{\mathbf{X}}}.\left( 2-2\left( \frac{\mathbf{X}-b}{a} \right) \right)&		b\leqslant \mathbf{X}<b+a\\
	0&		otherwise\\
\end{matrix} \right. ,\\
\end{aligned}
\end{equation}
where $b$ and $a$ represent the corresponding learnable bias and scale factor, respectively, and $L$ is the loss function. 

Due to the attention values (ranging from 0 to 1) significantly differ from the other activations, the attention is specifically binarized, as shown in Eq.~\ref{eqatt}.
\begin{equation}
\label{eqatt}
\begin{aligned}
	\mathbf{Forward}&~ \hat{\mathbf{A}}_{tt}=\mathrm{B}_{att}\left( \mathbf{A}_{tt},a,b \right)\! =\\
 &a \cdot clip\left( round\left( \frac{\mathbf{A}_{tt}-b}{a} \right) ,0,1 \right),\\
	\mathbf{Backward}&~\frac{\partial L}{\partial \mathbf{A}_{tt}}=\left\{ \begin{matrix}
	a\frac{\partial L}{\partial \hat{\mathbf{A}}_{tt}}&		b\leqslant \mathbf{A}_{tt}<a +b\\
	0&		otherwise\\
\end{matrix} \right. ,
\end{aligned}
\end{equation}
where $\mathrm{B}_{att}$ represents the binary function for the full precision attention matrix $\mathbf{A}_{tt}$, and $\hat{\mathbf{A}_{tt}}$ denotes the corresponding binary attention matrix. $clip\left( x,0,1 \right)$ truncates values that fall below 0 to 0 and those above 1 to 1, effectively ensuring that the output remains within the range [0, 1]. $round$ operation maps the input to the nearest integer.

To binarize weights, we adopt the commonly used operations in Eq.~\ref{eqw}.
\begin{equation}
\label{eqw}
\begin{aligned}
\mathbf{Forward}~&\mathrm{B}_w\left( \mathbf{W}_{\left[ :,k \right]} \right) =G\left( abs\left( \mathbf{W}_{\left[ :,k \right]} \right) \right) \\
&\cdot sign\left( \mathbf{W}_{\left[ :,k \right]} \right) ,
\\
\mathbf{Backward}~&\frac{\partial L}{\partial \mathbf{W}_{\left[ :,k \right]}}=G\left( abs\left( \mathbf{W}_{\left[ :,k \right]} \right) \right) \\
&\cdot \frac{\partial L}{\partial \mathbf{\hat{W}}_{\left[ :,k \right]}}\cdot \mathbf{1}_{-1<\mathbf{W}_{\left[ :,k \right]}<1},
\end{aligned}
\end{equation}
where $\mathbf{W}_{\left[ :,k \right]}$ represents the data in the $k$-th output channel of $\mathbf{W}$. $G\left( \right)$ represents the average function to calculate the scaling factor. $\mathbf{1}_{-1<\mathbf{W}_{\left[ :,k \right]}<1}$ denotes a mask tensor with the same size as $\mathbf{W}_{\left[ :,k \right]}$. The element of the mask tensor is marked as one if the corresponding element in $\mathbf{W}_{\left[ :,k \right]}$ falls in the closed interval [-1,1]. \\ 
\vspace{-10pt}
\begin{figure*}[htbp]
\setlength{\abovecaptionskip}{0pt}
\setlength{\belowcaptionskip}{-0pt}
\centering
\includegraphics[width=5.5in]{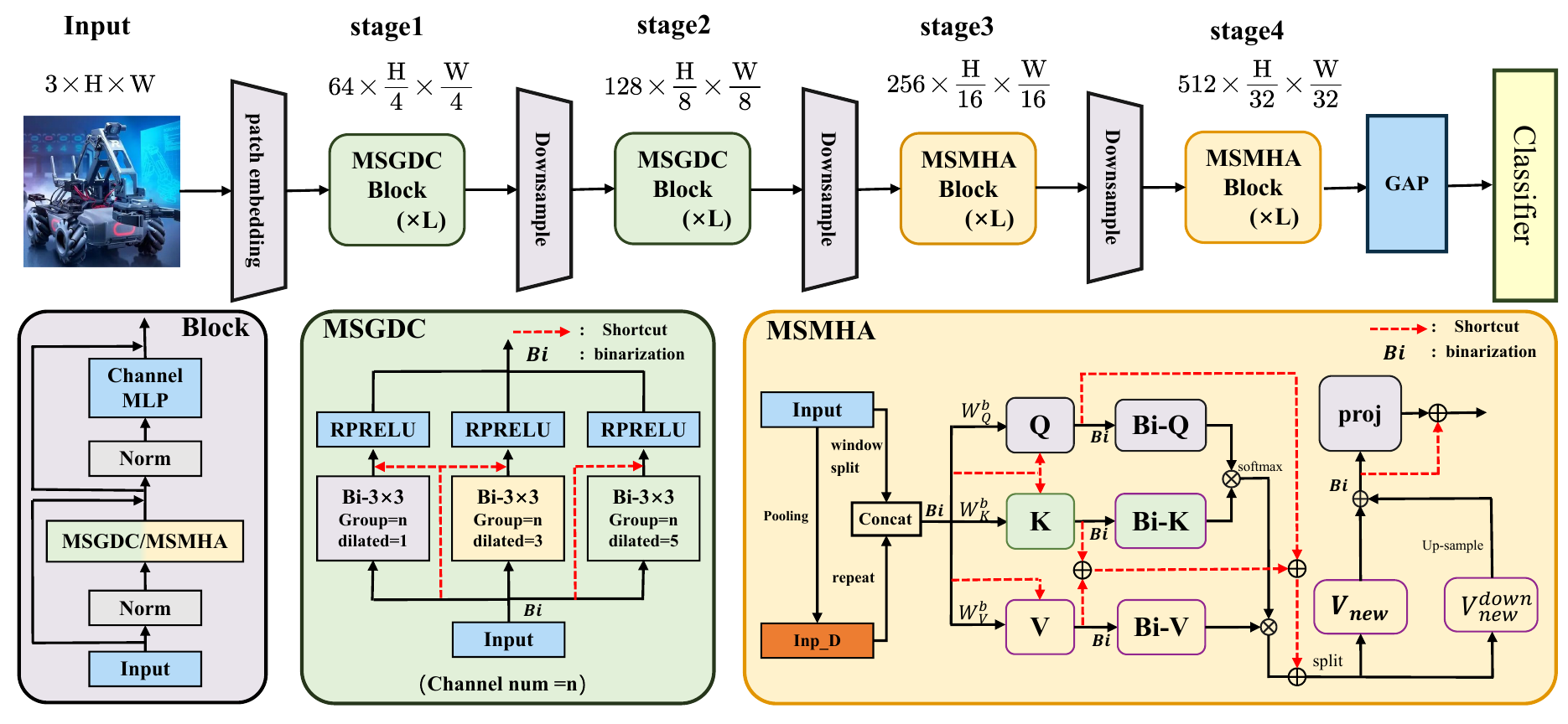}
\caption{The architecture of the proposed binary hybrid ViT. ``MSGDC" and ``MSMHA" refer to the Binary Multi-Scale Grouped Dilated Convolution module and the Binary Multi-Scale Multi-Head Attention module, respectively. ``GAP" stands for global average pooling. ``Input\_D" denotes the input tensor after downsampling. ``RPRELU" is the activation layer proposed in~\cite{liu2020reactnet}.}
\label{fig1}
\end{figure*}
\section{Method}
\label{Methods}
\subsection{Binarized Hybrid Vision Transformer}
The architecture of our approach is depicted in Fig.~\ref{fig1}, which comprises four stages in a feature pyramid with distinct feature sizes in spatial and channel dimensions in each stage. Given the input image $\mathbf{I}\in \mathbb{R} ^{3\times H\times W}$, the patch embedding layer based on the convolution, as shown in Eq.~\ref{eqpe}, is applied to split and project image $\mathbf{I}$ to the feature sequence $X_{0}\in \mathbb{R} ^{64\times \frac{H}{4}\times \frac{W}{4}}$, where $H$ and $W$ indicate horizontal and vertical size of image, respectively.
\begin{equation}
\label{eqpe}
\begin{aligned}
\mathbf{H}_0=\mathrm{GELU}\left( bn\left( Cov\left( \mathbf{I} \right) \right) \right) , \mathbf{X}_0=\mathbf{H}_0+\mathbf{P}_{\mathrm{e}},
\end{aligned}
\end{equation}
where $bn$ means a batch-norm layer. $\mathrm{GELU}$ is a $gelu$ activation function. $cov$ is a $4 \times 4$ convolution layer with a stride of 4. $\mathbf{P}_{\mathrm{e}}$ means a learnable position embedding. 

Then the embedding of $\mathbf{X}_{l-1}$ in each block as follow,
\begin{equation}
\label{eqblock}
\begin{aligned}
\mathbf{H}_{l-1}&=\left\{ \begin{array}{c}
	\!\!\mathrm{MSGDC}\left( \mathbf{X}_{l-1} \right) \!\!+\!\!\mathbf{X}_{l-1},if\,\,\!Stage\in \left[ 1,2 \right]\\
	\!\!\mathrm{MSMHA}\left( \mathbf{X}_{l-1} \right) \!\!+\!\!\mathbf{X}_{l-1},if\,\,\!Stage\in \left[ 3,4 \right]\\
\end{array} \right. \!\!\!\!,
\\
\mathbf{X}_l&=\mathrm{B}\_\mathrm{MLP}\left( \mathbf{H}_{l-1} \right) +\mathbf{H}_{l-1},
\end{aligned}
\end{equation}
For simplification, we use "MSGDC" and "MSMHA" to represent the Binary Multi-Scale Grouped Dilated Convolution module and the Binary Multi-Scale Multi-Head Attention module, respectively.

The pyramid structure can extract multi-scale features and effectively enhance the representation ability of binary features by increasing the channel dimension of the output feature. However, it also results in a large spatial resolution of features in the early stages of the model. It leads to a significant increase in the complexity of the attention module. To address this issue, we introduce MSGDC consisting of three $3\times3$ grouped atrous convolution layers (each layer employing a different dilated ratio) as the token mixer in the first two stages of the model. Meanwhile, we introduce an MSMHA module to accomplish token-wise feature fusion in the last two stages, ensuring effective information integration at different scales. A convolution layer with a stride of 2 and a $2 \times 2$ kernel size is applied in each down-sampling layer, which doubles the number of activation channels. 

\subsection{Token mixer}
\paragraph{Multi-scale grouped dilated convolution (MSGDC)} We apply three grouped convolutions with different dilation rates, enabling multi-scale feature fusion to enhance the representational capability of binary activation. Compared with ordinary convolutions and self-attention modules, grouped convolutions significantly reduce the model's parameters and computational complexity. 

For the input feature $\mathbf{X}_{l-1}\in \mathbb{R} ^{H\times W\times C}$, the embedding process of the MSGDC module is defined as
\begin{equation}
\label{eqMSGDC}
\begin{aligned}
\mathbf{H}_{l-1}^{n}&=\mathrm{RPReLU}\left(B\_Cov_{3\times 3,g}^{dil=2n-1}\left( B_a\left( \mathbf{X}_{l-1} \right) \right) + \mathbf{X}_{l-1}\right) ,\\ \mathbf{H}_{l-1}&=bn\left( \mathbf{H}_{l-1}^{1}+\mathbf{H}_{l-1}^{2}+\mathbf{H}_{l-1}^{3} \right),
\end{aligned}
\end{equation}
where $\mathrm{B}_{\mathrm{a}}\left(  \right) $ is the binary function defined in Eq.~\ref{eq2-1}. $B\_Cov_{3\times 3,g}^{dil=2n-1}$ is the binarized $3\times3$ grouped atrous convolution layer with dilated ratio of $2n-1$, $n\in \left( 1,2,3 \right) $. $\mathbf{H}_{l-1}^{1}$, $\mathbf{H}_{l-1}^{2}$, and $\mathbf{H}_{l-1}^{3}$ are the outputs of each grouped convolution layers, respectively. $\mathrm{RPReLU}$ is the activation function proposed by Liu et al.~\cite{liu2020reactnet}.

\paragraph{Multi-Scale Multi-Head Attention module (MSMHA)} The original self-attention mechanism requires calculating the similarity between all tokens. However, when $Q$ and $K$ are binarized, the distribution of attention values exhibits a long-tail distribution with nearly 99\% of attention values approaching 0~\cite{he2023bivit,gao2024gsb}, rendering most similarity calculations useless. To solve this problem, we introduce MSMHA, a variant of the window attention mechanism that maintains global information interaction and decreases computation costs. The motivation for applying the window attention mechanism is based on our observation that excessive numbers of tokens can degrade the performance of binary ViT, which is described in detail in \textbf{Observation}~\ref{lemma1} (The detailed illustration is shown in the Appendix.). 

\begin{figure}
\setlength{\belowcaptionskip}{-10pt}
\centering
\includegraphics[width=3.3in]{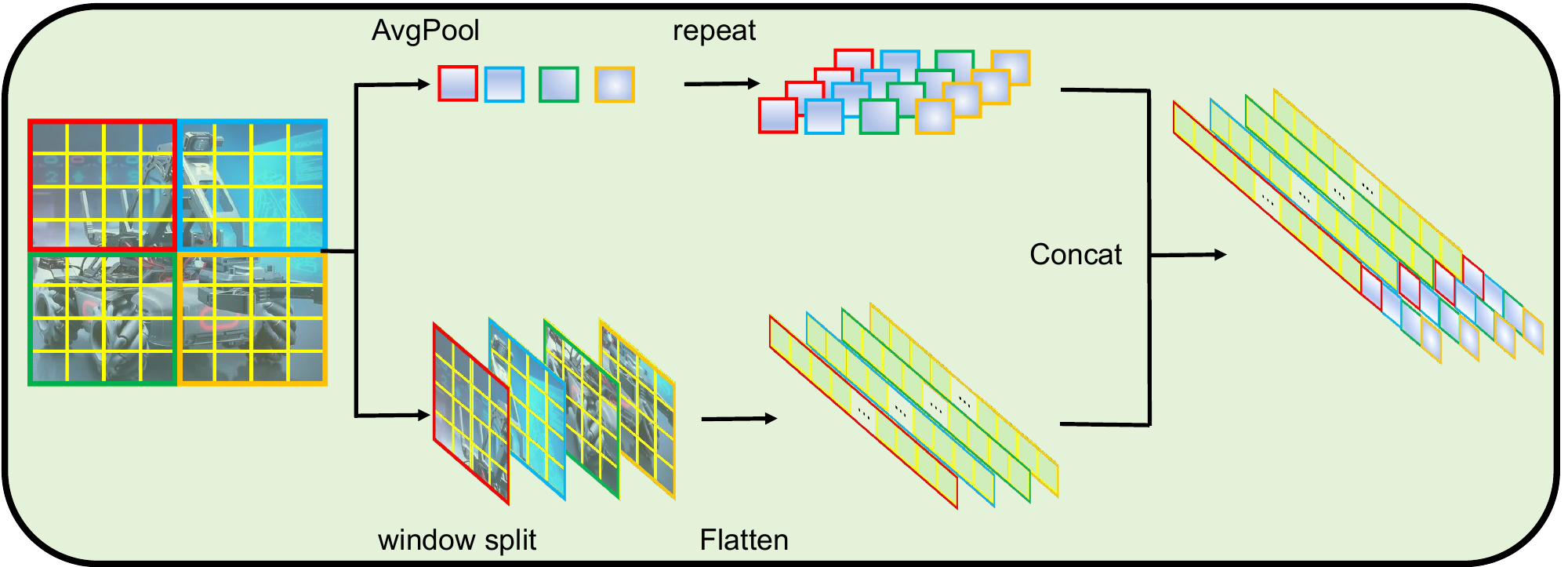}
\caption{Building hidden state feature for MSMHA.}
\label{fig4}
\end{figure}

\begin{Proposition}
\label{lemma1}
Avoiding excessive numbers of tokens is beneficial for Binary ViT.
\end{Proposition}
In MSMHA, for the input feature \small$\mathbf{X}_{l-1}\in \mathbb{R} ^{H\times W\times C}$\normalsize (Fig.~\ref{fig4}), we first apply average pooling with a kernel size of $7 \times 7$ to obtain high-scale feature $\mathbf{X}_{l-1}^{high}\in \mathbb{R} ^{\frac{H}{7}\times \frac{W}{7}\times C}$. Meanwhile, we split the spatial resolution of the input feature to $7 \times 7$ and get the window version of the input feature \small$\mathbf{X}_{l-1}^{win}\in \mathbb{R} ^{\frac{HW}{49}\times 7\times 7\times C}$\normalsize. Subsequently, the hidden state feature \small$\mathbf{H}\in \mathbb{R} ^{\frac{HW}{49}\times \left( 49+\frac{HW}{49} \right) \times C}$\normalsize is obtained by concatenating \small$\mathbf{X}_{l-1}^{win}$\normalsize (flattened to 1D vector) and \small$\mathbf{X}_{l-1}^{high}$\normalsize (repeated and flattened). As shown in Eq.~\ref{eqqkv2}, $\mathbf{Q}_{l-1}$, $\mathbf{K}_{l-1}$, and $\mathbf{V}_{l-1}$ tensors are obtained by applying three binary linear layers to the hidden state feature $\mathbf{H}$.
\begin{equation}
\label{eqqkv2}
\begin{aligned}
&\mathbf{Q}_{l-1}=\mathbf{B}_q\left( \mathbf{H} \right), \mathbf{K}_{l-1}=\mathbf{B}_k\left( \mathbf{H} \right), \mathbf{V}_{l-1}=\mathbf{B}_v\left( \mathbf{H} \right),
\\
&\mathbf{B}_q\left( \mathbf{X} \right) =\mathrm{RPReLU}\left( bn\left( \mathrm{B}_\mathrm{Lin}\left( \mathbf{X},W_q \right) \right) +\mathbf{X} \right) ,
\end{aligned}
\end{equation}
where $\mathbf{B}_q\left( \right)$, $\mathbf{B}_k\left( \right)$, and $\mathbf{B}_v\left( \right)$ are the same operation with three different tensors of binary weight. $\mathrm{B}_\mathrm{Lin}$ is a binary linear layer and $W_q$ is the binary weight of linear layer in $\mathbf{B}_q$. Then the attention matrix $\mathbf{A}_{tt}$ is obtained by Eq.~\ref{eqatt2}. 
\begin{equation}
\label{eqatt2}
\begin{aligned}
\mathbf{A}_{tt}=softmax\left( \frac{\mathbf{B}_a\left( \mathbf{Q}_{l-1} \right) *\mathbf{B}_a\left( \mathbf{K}_{l-1} \right)}{\sqrt{d}} \right).
\end{aligned}
\end{equation}
Note that the binary attention matrix employs 0 and 1 instead of -1 and 1 as the binary states. Unlike the full-precision attention matrix, the binary version cannot assign different weights to each token based on their similarity, which is one of the primary factors resulting in the significant performance drop in binary ViT. To address this issue, we propose an effective binarization method called Quantization Decomposition (QD).
\begin{figure}[t]
\setlength{\abovecaptionskip}{2pt}
\setlength{\belowcaptionskip}{-15pt}
\centering
\includegraphics[width=2.8in]{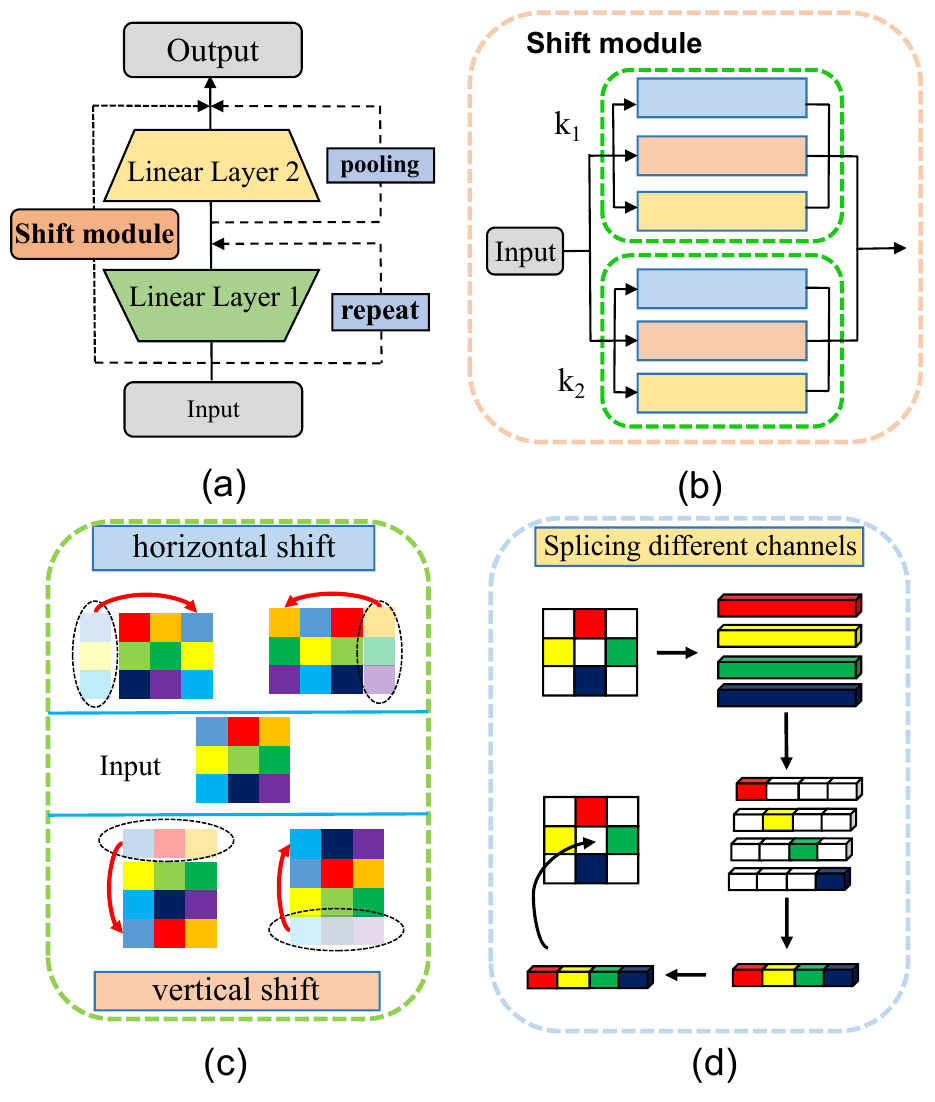}
\caption{Architecture of the binary MLP module. (a) Overall architecture. (b) The shift module, where colored rectangles represent horizontal, vertical, and mixed shift operations, with k1 and k2 denoting different translations. (c) and (d) illustrate specific shift operations.}
\label{fig3}
\end{figure}
\subsection{Quantization Decomposition}
The softmax function ensures that each element of the attention matrix is less than 1. Therefore, we introduce a global scaling constant $s=2^n-1$ (with $n=2$). The binarization of the attention matrix is then achieved through:
\begin{equation}
\label{eqbatt}
\begin{aligned}
\mathbf{\hat{A}}_{tt}^{\sigma}=\varphi \left( round\left( s\mathbf{A}_{tt} \right) \geqslant \sigma-0.5 \right) ,\sigma=\left( 1,2\cdots ,s \right) ,
\end{aligned}
\end{equation}
where $\varphi\left(\right)$ is a boolean function. In this way, we obtain $s$ binary attention matrices. If an element of $s\mathbf{A}_{tt}$ is larger than or equal to the constant $\sigma-0.5$, the corresponding element of $\mathbf{\hat{A}}_{tt}^{\sigma}$ is set to 1. In contrast to the original binarization function, QD involves a decomposition process consisting of logical operations with relatively low computational complexity.

Meanwhile, we observe that applying additional shortcut in each MSMHA can improve the optimization of the binary linear layer for $\mathbf{Q}$, $\mathbf{K}$, and $\mathbf{V}$ tensors in the attention module for better performance, which is detailed in \textbf{Observation}~\ref{lemma2} (The detailed illustration is shown in Appendix.). Different from the previous method~\cite{liu2018bi,le2023binaryvit} aiming to enhance the representational capacity of binary activations through residual connections, our focus is on alleviating optimization issues caused by gradient mismatch and gradient vanishing. Therefore, we add shortcuts from full-precision $\mathbf{Q}$, $\mathbf{K}$, and $\mathbf{V}$ tensors to the output of the attention module directly, which improves the optimization process by decreasing the gradient mismatch between each activation in the attention module and the module's output.
\begin{Proposition}
\label{lemma2}
Adding a residual connection in each binary layer is beneficial for Binary ViT.
\end{Proposition}
 
After obtaining multiple binarized attention matrices, $\mathbf{V}_l$ can be obtained through Eq.~\ref{eqvn}. 
\begin{equation}
\label{eqvn}
\begin{aligned}
\mathbf{V}_l=\left( \sum_{\sigma=1}^s{\mathbf{\hat{A}}_{tt}^{\sigma}}\circledast \mathbf{B}_a\left( \mathbf{V}_{l-1} \right) \right) +\mathbf{Q}_{l-1}+\mathbf{K}_{l-1}+\mathbf{V}_{l-1},
\end{aligned}
\end{equation}

According to the recorded splicing dimension information (the `concatenation' process shown in Fig.~\ref{fig4}) in obtaining the hidden state feature $\mathbf{H}$, $\mathbf{V}_l$ is split to the window-version feature $\mathbf{V}_{l}^{win}$ and the high-scale feature $ \mathbf{V}_{l}^{high}$. Finally, the output feature $\mathbf{X}_l$ is obtained by Eq.~\ref{eqxl}.
\begin{equation}
\label{eqxl}
\begin{aligned}
\mathbf{X}_l=\vartheta\left( \mathbf{V}_{l}^{win} \right) +\Uparrow \left( mean\left(\mathbf{V}_{l}^{down} \right)\right),
\end{aligned}
\end{equation}
where $\vartheta$ refers to reshaping the window-version feature to the original feature. $\Uparrow $ is the nearest neighbor interpolation to make the shape of $\mathbf{V}_{l}^{down}$ the same as $\mathbf{X}_l$. $mean\left( \mathbf{X}\right) $ is a global average function along the dimension corresponding to the number of windows.

\subsection{Binary MLP}
The architecture of our proposed binary MLP module is depicted in Fig.~\ref{fig3} (a). To mitigate information loss and gradient errors, we introduce the shift operation, as shown in Fig.~\ref{fig3} (b), (c), and (d). 

Given the input feature $\mathbf{H}_{l-1}\in \mathbb{R} ^{H\times W\times C}$, the processing of the binary MLP is defined as
\vspace{-1pt}
\begin{equation}
\label{eqbffn}
\begin{aligned}
&\mathbf{H}_{l-1}^{10}=bn\left( \mathrm{B}_{\mathrm{L}1}\left( \mathrm{B}_a\left( \mathbf{H}_{l-1} \right) \right) \right) ,
\\
&\mathbf{H}_{l-1}^{1}=\mathrm{RPReLU}\left( \mathbf{H}_{l-1}^{10}+repeat\left( \mathbf{H}_{l-1} \right) \right) ,
\\
&\mathbf{H}_{l-1}^{20}=bn\left( \mathrm{B}_{\mathrm{L}2}\left( \mathrm{B}_a\left( \mathbf{H}_{l-1}^{1} \right) \right) \right) ,
\\
&\mathbf{H}_{l-1}^{2}=\mathrm{RPReLU}\left( \mathbf{H}_{l-1}^{20}+pool\left( \mathbf{H}_{l-1}^{1} \right) \right) ,
\\
&\mathbf{S}_k=L_s\left( \mathbf{S}_{k}^{hor}\left( \mathbf{H}_{l-1} \right) \right) +L_s\left( \mathbf{S}_{k}^{ver}\left( \mathbf{H}_{l-1} \right) \right) 
\\
&\qquad +L_s\left( \mathbf{S}_{k}^{mix}\left( \mathbf{H}_{l-1} \right) \right) ,k\in \left( 1,2 \right) ,
\\
&\mathbf{X}_l=\mathbf{H}_{l-1}^{2}+\mathbf{S}_1+\mathbf{S}_2,
\end{aligned}
\end{equation}
where $\mathrm{B}_\mathrm{L1}$ and $\mathrm{B}_\mathrm{L2}$ are two binary linear layers with a ratio of 4 and 0.25 between the number of input channels and output channels, respectively. $L_s$ denotes a learnable channel-wise scaling transformation. The $repeat$ operation involves repeating the input four times and concatenating them along the channel dimension. The $pool$ function denotes 1D average pooling with a stride of 4.
$\mathbf{S}_1$ and $\mathbf{S}_2$ represent two groups of shift operations with different strides. $\mathbf{S}^{hor}$, $\mathbf{S}^{ver}$, and $\mathbf{S}^{mix}$ are three fundamental shift operations defined in Fig.~\ref{fig3} (c) and (d), respectively.

For the horizontal shift operation shown in Fig.~\ref{fig3} (c), shifting the entire input feature map one pixel to the right is equivalent to moving the first column of the input feature map to the last column. Vertical shift operations can be implemented similarly. For the mix shift operation shown in Fig.~\ref{fig3} (d), we first identify four neighboring tokens of the current token. Then, in sequential order, we extract features from each adjacent token by taking one-fourth of the features along the channel dimension and concatenating them to replace the features of the current token.

\subsection{Training Settings}
\noindent\textbf{Distillation}~~Compared to the one hot label, the prediction provided by the teacher model for each class contains more information. Therefore, we utilize the DeiT-small~\cite{touvron2021training} model as the teacher model during the training process to enhance the performance of the proposed binary model.

Compared with real-valued networks, weight oscillation occurs more frequently in the binary model~\cite{nagel2022overcoming,liu2023oscillation,xu2023resilient}. In addition, we find that the commonly used Adam optimizer enlarges the weight oscillation of binary networks and results in some weights eventually ceasing updating, as described in Observation.~\ref{lemma3} (The detailed illustration is shown in Appendix.). 
\begin{Proposition}
\label{lemma3}
The Adam optimizer enlarges the weight oscillation of binary networks in the later stages of the training process, failing to update numerous parameters effectively.
\end{Proposition}
\noindent To tackle this issue, we integrate L$_1$-regularization into the model's latent weights during the later phases of the model training, thereby compelling some latent weights around 0 close to ±1. Then, the total loss function is shown in Eq.~\ref{eqloss}.
\begin{equation}
\label{eqloss}
\begin{aligned}
L&=\left( 1-\lambda -\beta \right) L_{cls}+\lambda L_{dis}+\beta L_{re},
\\
L_{re}&=\frac{1}{n}\sum_{i=1}^n{\left| \left| w_i \right|-1 \right|}, \\
\beta &=\left\{ \begin{matrix}
	0.1&		if\,\,T_{now}\geqslant 0.9 \times T_{\max}\\
	0&		others\\
\end{matrix} \right.,
\end{aligned}
\end{equation}
where $L_{cls}$ and $L_{dis}$ represent the cross-entropy loss between the class prediction and the corresponding label and the one between the output of the teacher network and the student networks' output, respectively. $L_{re}$ refers to the L$_1$-regularization loss. To balance each component of $L$, we set the coefficient $\beta = 0.1$ ($L_{re}$ is about 10 times of $L_{cls}$ and $L_{dis}$ in scale). Additionally, the model achieves the best performance when the hyperparameter $\lambda$ is set to 0.8. $T_{now}$ and $T_{max}$ mean the current epoch index and the number of total epochs, respectively. \\
\section{Experiments}
\subsection{Datasets and Implementation Details}
\textbf{Datasets.}~We evaluate the proposed method on four datasets, CIFAR-10~\cite{krizhevsky2009learning}, ImageNet-1K~\cite{deng2009imagenet}, ADE20K~\cite{zhou2019semantic} and the RS-LVF Dataset~\cite{yin2024pathfinder}. 
CIFAR-10 consists of 60,000 images with ten classes, where 50,000 images are used for training and 10,000 for testing. ImageNet-1K has 1,000 classes, with a training set of 1.2 million images and a test set of 50,000. ADE20K is a challenging dataset including more than 20000 images with 150 categories with a limited amount of training data per class. RS-LVF comprises 1000 aerial images with corresponding road labels in a Bird's-Eye View (BEV).\\
\textbf{Implementation Details.}~We apply "RandomResizedCrop" and "RandomHorizontalFlip" operations to the ImageNet-1K dataset for data augmentation. For the CIFAR-10 dataset, we follow the data augmentation scheme proposed in DeiT~\cite{touvron2021training}. The AdamW optimizer with cosine annealing learning-rate decay and an initial learning rate of $5\times 10^{-4}$ are applied to train the proposed method. The training process for the ImageNet-1K dataset is conducted using 4 NVIDIA A100 GPUs with a batch size of 512. The total epoch number for the ImagNet-1k dataset and the CIFAR-10 dataset are 150 and 300, respectively.  For ADE20K and RS-LVF, the training batch sizes are 18 and 4, and the epochs are 50 and 100, respectively.
\subsection{Classification}
\textbf{Results On CIFAR-10.}~We compare the classification accuracy of our method on the CIFAR-10 dataset with other BNN and binary ViT models. The result is shown in Tab.~\ref{tab2}. Since the ViT model requires a large amount of training data~\cite{gao2024gsb} and lacks the inductive bias specific to visual tasks, the performance of binary ViT methods on small datasets is typically weaker than the BNN methods. Nevertheless, compared to BNN methods, our proposed BHViT-Tiny method achieves higher classification accuracy (0.5\%) than the SOTA method ReCU~\cite{xu2021ReCU}. For the methods based on ViT architecture, with the same level of the parameter's number, our proposed BHViT-Small outperforms ProxConnect++$\ddagger$~\cite{lu2023understanding} by nearly 8.81\% and GSB~\cite{gao2024gsb} by 4.8\% in terms of accuracy.
\begin{table}[t]
\setlength{\abovecaptionskip}{0pt}
\setlength{\belowcaptionskip}{0pt}
\caption{Classification results on CIFAR-10. W-A refers to the bit number of weights and activations for the corresponding binary method. NP denotes the number of network parameters ($\times 10^6$). The $\ddagger$ indicates that the attention matrix is fully precise.}
		\label{tab2}
		\centering
  \small
		\begin{tabular}{ccccc}  %确定表格竖行格式   设置列宽度：将c换为p{ cm}
			\Xhline{1.5pt}
			 Architecture   &Methods                        & W-A     & NP       &  Top-1(\%)           \\
			\Xhline{1.0pt}	
			\multirow{8}*{ResNet-18}&Bi-RealNet~\cite{liu2018bi}      & 1-1     &11.2         & 89.12                \\
            
           ~& IR-Net~\cite{qin2020forward}                             & 1-1    &11.2            & 91.20                  \\
            
           ~& RBNN~\cite{lin2020rotated}                             & 1-1     &11.2            & 92.20                  \\
            
			~&XNOR-Net~\cite{rastegari2016xnor}                       & 1-1   &11.2             & 90.21                   \\
            
			~& RAD~\cite{ding2019regularizing}                              & 1-1  &11.2               & 90.05                 \\
            ~& Proxy-BNN~\cite{he2020proxybnn}                              & 1-1    &11.2              & 91.80                 \\
           ~& ReActNet~\cite{liu2020reactnet}                              & 1-1    &11.2              & 92.31                  \\
          ~&  ReCU~\cite{xu2021ReCU}                                      & 1-1     &11.2               & 92.80                  \\
           \hline
          Mobile-Net&  ReActNet-A~\cite{liu2020reactnet}                               & 1-1   &28.3           & 82.95                    \\
            \hline
         
         \multirow{2}*{DeiT-Small} &  ProxC++$\ddagger$~\cite{lu2023understanding}                              & 1-1 &21.6             & 86.19                    \\
          &  GSB~\cite{gao2024gsb}                             & 1-1 &21.6             & 91.20                   \\
            \hline
		\multirow{2}*{BHViT}&	Ours-Tiny                                              & 1-1   &13.2              &  93.30                   \\
        
		~&	Ours-Small                                                      & 1-1          &22.1        &   \textbf{95.00}      \\
		\Xhline{1.5pt}
		\end{tabular}
	\end{table}
\begin{figure}[t]
\centering
\includegraphics[width=3.3in]{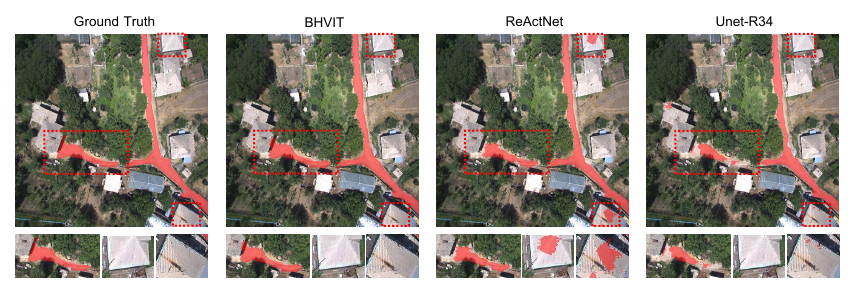}
\caption{Visualization on the RS-LVF dataset.}
\label{figrslvf}
\end{figure} 

\noindent\textbf{Results On ImageNet-1K.}~On the ImageNet-1K dataset, the classification performance of binary models based on different architectures is shown in Tab.~\ref{tab3}. We find the methods with full-precision downsampling layers generally achieve good performance. Compared with ResNet-18 based methods, the performance of BHViT-tiny$^{\dagger}$ is higher than the best method, ReActNet, by about 0.5\% with similar OPs and model size. From Tab.~\ref{tab3}, we can see the model's performance based on pure ViT architecture is not good enough. BHViT-Small$^{\dagger}$ is 20.6 \% higher than the current SOTA (ReActNet) with a similar OPs level. Compared with the Swin transformer architecture, BHViT-Small$^{\dagger}$ exceeds the best method (BiViT) by 11.5 \%. For the BinaryViT architecture, our method BHViT-Small still exceeds it by 0.7 \% with the same setting. Compared with ReActNet-B*, based on the MobileNet architecture, our BHViT-Small$^{\dagger}$ achieves the same performance with a smaller model size and computational cost. In addition, the performance of ReActNet-A method on the datasets with relatively limited training data (e.g. CIFAR-10) is far inferior to the proposed algorithm. The proposed algorithm solves the problem shown in Fig.~\ref{figduibik} and greatly surpasses current transformer-based algorithms in terms of performance, enabling the binary ViT architecture to outperform binary CNNs.

\subsection{Road Segmentation and Image Segmentation} 
\noindent \textbf{Road Segmentation.} We evaluate the performance of the proposed model in road segmentation with the BEV perspective in aerial images. Road segmentation can be considered as a pixel-wise binary classification task. To assess the method's performance, we utilize the Mean Intersection over Union (mIOU) metric and mean accuracy metrics. The result is shown in Tab.~\ref{Tabrslvf}, and some visual results for each method are demonstrated in Fig.~\ref{figrslvf}. Each method is based on an encoder-decoder architecture, similar to U-Net~\cite{ronneberger2015u}. The decoder module for each method is a ResNet structure with increasing resolution. In particular, different from ReActNet~\cite{liu2020reactnet} and our method, the weight and activation of the ResNet-34 method are kept in full precision. From Tab.~\ref{Tabrslvf}, we can observe that our proposed BHViT model exhibits higher segmentation accuracy than the CNN-based ReActNet, surpassing even full-precision methods.

\noindent \textbf{Image Segmentation.} We evaluate the performance of the proposed model in image segmentation with ADE20K~\cite{zhou2019semantic} dataset. We apply two evaluation metrics, including pixel accuracy (pixAcc) and mean Intersection-over-Union (mIoU) to test each method. The result is shown in Tab~\ref{tabade20k}, which demonstrates that our BHViT achieves SOTA performance among current binary segmentation algorithms.
\begin{table}[htbp]
\setlength{\abovecaptionskip}{0pt}
\setlength{\belowcaptionskip}{0pt}
    	\caption{Results On ImageNet-1K. $\dagger$ represents keeping the downsampling layer at full precision and $*$ means applying the two stages training scheme proposed by~\cite{martinez2020training}. $\mathrm{OPs}$ is defined as $\mathrm{OPs}=\frac{\mathrm{BOPs}}{64}+\mathrm{FLOPs}$~\cite{liu2020reactnet}. BOPs and FLOPs mean binary and float operations, respectively.}
		\label{tab3}
		\centering
            \setlength{\tabcolsep}{2.5pt}{
		\begin{tabular}{ccccc}  %确定表格竖行格式   设置列宽度：将c换为p{ cm}
		\Xhline{1.5pt}
		\multirow{2}*{Network}	     &\multirow{2}*{Methods}            & Size        &  OPs &  Top-1                \\
       &            & (MB)        &  ($\times 10^8$)  &  (\%)                \\
		\Xhline{1.0pt}
		\multirow{9}*{ResNet-18~\cite{he2016deep}}	 &Real-valued        &  46.8        &  18.2  &  69.6               \\
            \cdashline{2-5}
             ~	 &BNNs$^{\dagger}$~\cite{courbariaux2016binarized}                  & \multirow{8}*{2.1}         &  \multirow{8}*{1.6}  &  42.2                 \\
             ~	 &XNOR-Net$^{\dagger}$~\cite{rastegari2016xnor}           &             &    &  51.2                \\
             ~	 &ABC-Net$^{\dagger}$ ~\cite{lin2017towards}           &               &    &  42.7                 \\
             ~	 &Bi-Real Net$^{\dagger}$~\cite{liu2018bi}        &              &    &  56.4                \\
             ~	 &IR-Net$^{\dagger}$~\cite{qin2020forward}             &               &    &  58.1                 \\
             ~	 &RBNN$^{\dagger}$~\cite{lin2020rotated}               &             &    &  59.6                 \\
             ~	 &ReCU$^{\dagger}$~\cite{xu2021ReCU}               &             &    &  61.0                \\
             ~	 &ReActNet*$^{\dagger}$~\cite{liu2020reactnet}                &              &    &  65.5                 \\
            \hline
            \multirow{5}*{DeiT-Tiny~\cite{touvron2021training}}	 &Real-valued        &  22.8 &   12.3      &  72.2                \\
            \cdashline{2-5}
            ~	 &BiT~\cite{liu2022bit}                   & \multirow{4}*{1.0}         &  \multirow{4}*{0.6}   &  21.7               \\
            ~	 &Bi-ViT~\cite{li2024bi}             &            &   &  28.7                 \\
            ~	 &BiBERT~\cite{qin2022bibert}             &           &    &  5.9                  \\
            ~	 &BVT-IMA~\cite{wang2024bvt}           &             &     &  30.03              \\
            \hline
            \multirow{3}*{\textbf{Ours-Tiny}}    &Real-valued & 54.8      &  22.4  &     78.5          \\
             \cdashline{2-5}
               &BHViT-Tiny   &  \multirow{2}*{2.5}      &  0.5  &    64.0              \\
           
               &BHViT-Tiny$^{\dagger}$                &        &  1.1   &  66.0              \\
           \Xhline{1pt}
            \multirow{4}*{ResNet-34~\cite{he2016deep}}	 &Real-valued          & 87.2       &  36.8  &  73.3              \\
            \cdashline{2-5}
             ~	 &Bi-Real Net$^{\dagger}$                &  \multirow{3}*{3.3}         &  \multirow{3}*{1.9}  &  62.2                 \\
             ~	 &RBNN$^{\dagger}$           &              &    &  63.1                 \\
             ~	 &ReCU$^{\dagger}$            &               &    &  65.1                \\
            \hline
             \multirow{2}*{MobileNet~\cite{howard2017mobilenets}}	 &ReActNet-A*                 &  \multirow{2}*{3.7}        &  0.87  &  69.4                 \\
             ~	 &ReActNet-B*$^{\dagger}$           &              & 1.63   &  70.1                 \\
            \hline         
            \multirow{6}*{DeiT-Small~\cite{touvron2021training}}	 &Real-valued        & 88.4         &  45.5  &  79.9                \\
            \cdashline{2-5}
            ~	 &BiT               & \multirow{5}*{3.4}         &   \multirow{5}*{1.5}  &  30.74                 \\
            ~	 &BVT-IMA        &            &   &  47.98                \\
            ~	 &Bi-ViT        &             &    &  40.9                \\
            ~	 &BiBERT        &             &    &  17.4              \\
            ~	 &ReActNet       &             &    &  49.5             \\
            ~	 &Si-BiViT~\cite{yin2024si}               & 4.6       &   2.2  &  55.67\\
            \hline
              \multirow{2}*{BinaryViT~\cite{le2023binaryvit}}	 &Real-valued         & 90.4         &  46.0  &  79.9                \\
            \cdashline{2-5}
            ~	 &BinaryViT               & \multirow{1}*{3.5}         &   \multirow{1}*{0.79}  &  67.7                \\
            \hline
            \multirow{4}*{Swin-Tiny~\cite{liu2021swin}}	 &Real-valued          & 114.2       &  44.9  &  81.2                \\
            \cdashline{2-5}
            ~	 &BiBERT          & \multirow{3}*{4.2}        & \multirow{3}*{1.5}  &  34.0                \\
            ~	 &Bi-ViT         &            &    &  55.5                 \\
            ~	 &BiViT~\cite{he2023bivit}         &            &    &  58.6                 \\
            ~	 &Si-BiViT         & 9.87        & 4.5  &  63.8               \\
              \hline
           
           \multirow{3}*{\textbf{Ours-Small}}    &Real-valued   & 90.4      &  45.0  &     79.3          \\
             \cdashline{2-5}
               &BHViT-Small   &  \multirow{2}*{3.5}      &  0.8   &    68.4              \\
           
               &BHViT-Small$^{\dagger}$               &        &  1.5   &  \textbf{70.1}             \\
		\Xhline{1.5pt}
		\end{tabular}}
	\end{table}
\subsection{Ablation Study}
\textbf{The impact of each proposed module.} As shown in Tab.~\ref{tabxr}, the result displays the performance changes of the BHViT-Small model on the CIFAR-10 dataset by sequentially dropping the proposed modules, validating each proposed module's effectiveness. In the Tab.~\ref{tabxr}, 'FDL' denotes the downsampling layer with full precision. 'RL' means the proposed regularization loss. 'Shift' is the proposed shift module shown in Fig.~\ref{fig3}. In particular, the ablation of 'MSGDC' and 'MSMHA' represents whether it adds multi-scale information interaction in the corresponding module.
\begin{table}[h]
\setlength{\abovecaptionskip}{0pt}
\setlength{\belowcaptionskip}{0pt}
\centering
\caption{Comparison results of road segmentation in aerial view.}
\scalebox{0.88}{
\begin{tabular}{cccccc}
\Xhline{1.5pt}
Encoder  & Architecture & W-A & OPs(G) & mAcc & mIou   \\
\Xhline{1.0pt}
ResNet-34 &U-Net &32-32 & 41.82 & 85.4 & 77.8      \\
ReActNet &U-Net &1-1  & 4.87  & 76.5 & 63.6      \\
Ours &U-Net &1-1 &  4.82 & \textbf{92.2} & \textbf{85.1}     \\
\Xhline{1.5pt}
\end{tabular}}
\label{Tabrslvf}
\end{table}
\begin{table}[htbp]
\setlength{\abovecaptionskip}{0pt}
\setlength{\belowcaptionskip}{0pt}
\setlength\tabcolsep{4pt}
\renewcommand\arraystretch{1.0}
\caption{Image segmentation results on ADE20K.}
\label{tabade20k}
\centering
		\begin{tabular}{ccccc}  %确定表格竖行格式   设置列宽度：将c换为p{ cm}
			\Xhline{1.5pt}
			   Method & Bit  & OPs (G)& pixAcc (\%) & mIoU  (\%) \\
			\Xhline{1.0pt}	
			  BNN & 1 & 4.84 & 61.69 & 8.68\\    
            \cdashline{1-5}
              ReActNet & 1 & 4.98 & 62.77 & 9.22\\     
              \cdashline{1-5}       
              AdaBin~\cite{tu2022adabin} & 1 & 5.24 & 59.47& 7.16\\   
             \cdashline{1-5}
			 BiSRNet~\cite{ding2022bi} & 1 & 5.07 & 62.85 & 9.74\\  
             \cdashline{1-5}
               Ours & 1 & 4.95 & \textbf{65.63} & \textbf{14.87}\\  
		\Xhline{1.5pt}
		\end{tabular}
\end{table} 
\begin{table}[htbp]
\setlength{\abovecaptionskip}{0pt}
\setlength{\belowcaptionskip}{0pt}
\setlength\tabcolsep{3pt}
\renewcommand\arraystretch{1.0}
\caption{Ablation study for BHViT-Small on the CIFAR-10.}
\label{tabxr}
\centering
		\begin{tabular}{ccccccc}  %确定表格竖行格式   设置列宽度：将c换为p{ cm}
			\Xhline{1.5pt}
			   Shift & MSGDC & MSMHA & QD & RL & FDL& Top1 (\%) \\
			\Xhline{1.0pt}	
			  $\checkmark$&$\checkmark$&$\checkmark$&$\checkmark$&$\checkmark$&$\checkmark$& 95.0\\    
            \cdashline{1-7}
              $\checkmark$&$\checkmark$&$\checkmark$&$\checkmark$&$\checkmark$&& 92.1 \\     \cdashline{1-7}       
              $\checkmark$&$\checkmark$&$\checkmark$&$\checkmark$&&&90.7  \\ 
             \cdashline{1-7}
			  $\checkmark$&$\checkmark$&$\checkmark$&&&& 88.9 \\ 
             \cdashline{1-7}
               $\checkmark$&$\checkmark$&&&&&86.7  \\ 
              \cdashline{1-7}
			 $\checkmark$&&&&&&85.6 \\ 
              \cdashline{1-7}
             &&&&&&83.2  \\ 
		\Xhline{1.5pt}
		\end{tabular}
\end{table} 

\noindent\textbf{The impact of coefficient $\lambda$.} Based on the CIFAR-10 dataset, we examine the influence introduced by the balance parameters $\lambda$ on the classification performance of the BHViT-Small model. The result is shown in Tab.~\ref{tablamda}, from which we can conclude that the model achieves the highest accuracy when the hyperparameter $\lambda$ is set to 0.8. 
\begin{table}[htbp]
\setlength\tabcolsep{2pt}
\setlength{\abovecaptionskip}{0pt}
\setlength{\belowcaptionskip}{0pt}
\renewcommand\arraystretch{1.0}
    	\caption{Ablation study for BHViT-Small on the CIFAR-10 dataset with different hyperparameters $\lambda$. }
		\label{tablamda}
		\centering
		\begin{tabular}{cccccccccc}  
			\Xhline{1.5pt}
			  $\lambda$&  0.1  & 0.2 & 0.3 & 0.4 & 0.5 & 0.6 &0.7& 0.8&0.9 \\
			\Xhline{1.0pt}	
			  Top1 (\%)& 93.6   & 93.8 &  94.1 & 94.2  & 93.9 &  94.5 & 94.8 & 95.0 &94.7\\    
		\Xhline{1.5pt}
		\end{tabular}
	\end{table}
\begin{table}[htbp]
\setlength{\abovecaptionskip}{0pt}
\setlength{\belowcaptionskip}{0pt}
\renewcommand\arraystretch{1.0}
    	\caption{Classification results of BHViT with different token mixer settings. $\dagger$ means the downsampling layer is maintained at full precision.}
		\label{purevit}
		\centering
         \setlength{\tabcolsep}{15pt}{
		\begin{tabular}{ccc}  
			\Xhline{1pt}
			  Model&Token mixer & Top1(\%) \\
			\Xhline{1pt}	
     BHViT-Small$^{\dagger}$ & Hybrid& 70.1 \\
     BHViT-Small$^{\dagger}$ & MSMHA & 68.8 \\
     BHViT-Small$^{\dagger}$ & MSGDC & 67.2 \\
		\Xhline{1pt}
		\end{tabular}}
	\end{table}
 
\begin{figure}[t]
\setlength{\abovecaptionskip}{0pt}
\setlength{\belowcaptionskip}{0pt}
\centering
\includegraphics[width=2.2in]{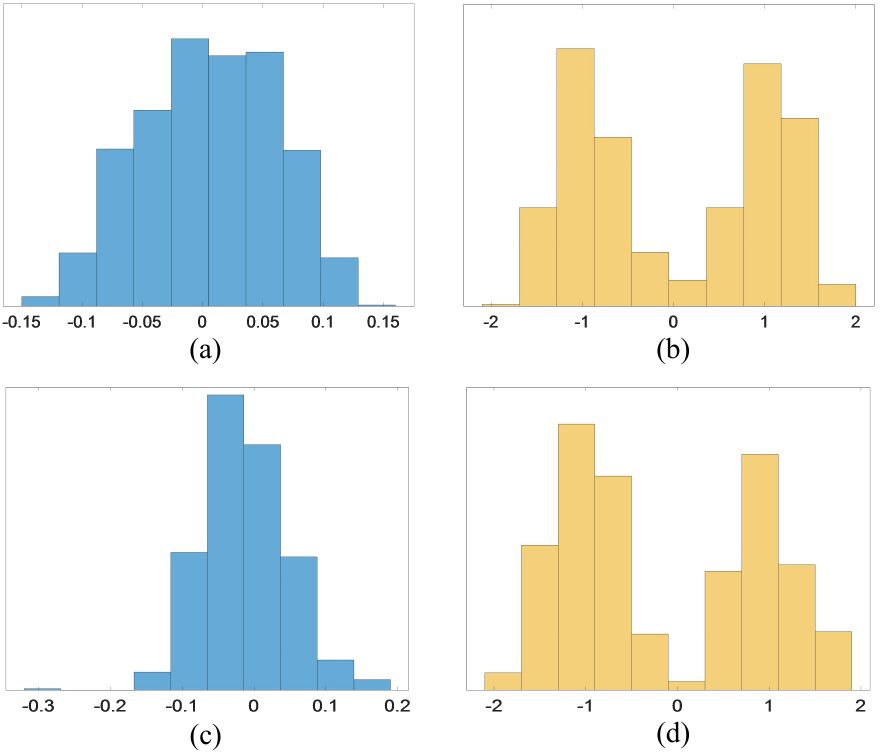}
\caption{Compare the weight histograms of two layers in different blocks. (a) and (c) denote the weight distribution of two layers without regularization loss. (b) and (d) denote the weight distribution of two layers with regularization loss.}
\label{figdc}
\end{figure}

\begin{figure}[t]
\setlength{\abovecaptionskip}{0pt}
\setlength{\belowcaptionskip}{0pt}
\centering
\includegraphics[width=3.0in]{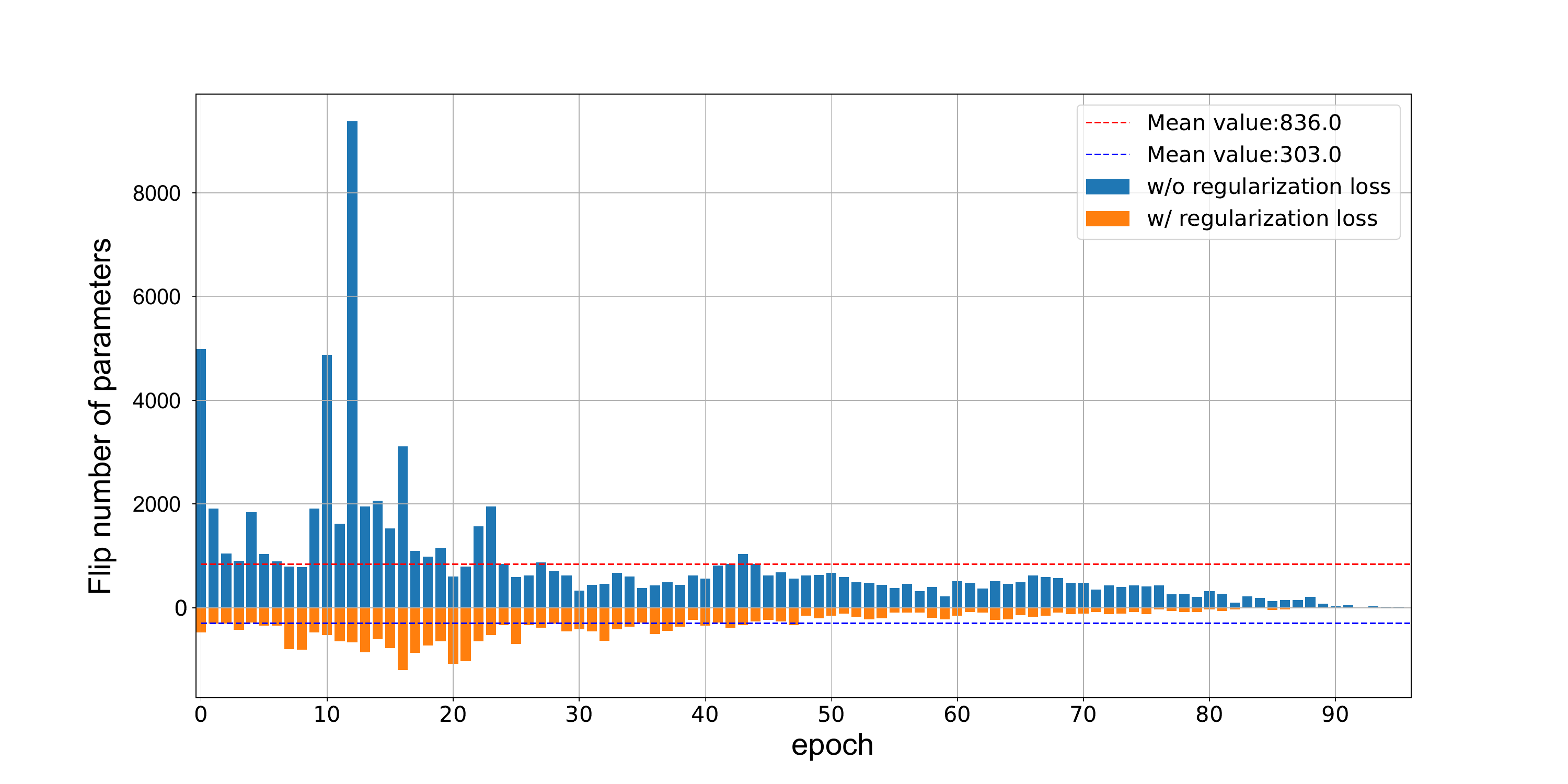}
\caption{The number of flipped weights w/wo RL.}
\label{figA}
\end{figure}

\noindent\textbf{The impact of Regularization Loss (RL).} We validate the effectiveness of the proposed regularization loss function from the perspective of weight distribution. The ablation of the RL on the CIFAR-10 dataset are depicted in Fig.~\ref{figdc} (The first convolution layer of block 1 and the $\mathrm{Q}$ linear layer of block 8, respectively), from which we observe that the RL change the distribution of latent weights, closer to +1 or -1, effectively mitigating weight oscillation. Meanwhile, at each training epoch, Fig.~\ref{figA} shows the change in the number of flipped parameters with or without RL (the first convolution layer of block 1), validating the effectiveness of RL in solving weight oscillation.

\noindent\textbf{The impact of different token mixer.}
We further studied the impact of different token mixer settings on model performance. The proposed MSMHA module can also reduce the number of tokens in the first two stages, i.e., applying MSMHA as the token mixer to all stages will turn the model into a pure ViT architecture. Alternatively, using MSGDC as the token mixer for each stage results in a pure CNN architecture. We validate the three different types of token mixers on ImageNet-1K, shown in Tab.~\ref{purevit}. The pure ViT version achieves 1.3\% lower classification accuracy than the hybrid version, while the pure CNN version performs the worst. The proposed hybrid architecture is a better choice for binarization.

 \section{Conclusions}
 In this paper, we propose a high-performance hybrid ViT framework and its binarized version, significantly reducing computational complexity while maintaining exceptional accuracy. Our work is based on three important observations, which guide us in introducing MSMHA, MSGDC, and MLP enhancement modules to improve the performance of binary ViT. Additionally, we propose the QD binarization method for the attention matrix and a regularization loss function to address weight oscillation issues in binary models when using the Adam optimizer. Experimental results show that the proposed method achieves SOTA performance among binarized models on benchmark datasets.
 
\noindent \textbf{Acknowledgements.} This work was supported by the fund from the Fundo para o Desenvolvimento das Ciências e da Tecnologia (FDCT) of Macao SAR with Reference No. 0067/2023/AFJ.
{
    \small
    \bibliographystyle{ieeenat_fullname}
    \bibliography{main}
}
%\maketitle
\newpage
\maketitlesupplementary
\setcounter{table}{0}
\setcounter{figure}{0}
\setcounter{section}{0}
\setcounter{equation}{0}
\setcounter{Proposition}{0}
\section{Hybrid Vision Transformer}
\label{3-1}
Recent advancements in ViT architectures have explored the integration of convolutional layers, departing from the original design that relied solely on fully connected layers for processing. Notably, works like Pyramid Vision Transformer (PVT) \cite{wang2021pyramid} and FastViT \cite{vasu2023fastvit} have introduced convolutional layers into the ViT model, leading to enhanced model performance and capabilities. Moreover, according to Meta-former~\cite{yu2022metaformer}, the outstanding performance of ViT is attributed more to its architectural characteristics rather than the introduction of self-attention modules. This finding further reinforces the effectiveness of the hybrid ViT architecture. Therefore, when designing model architectures based on ViT, using different structures as alternatives to the self-attention module for token mixing in specific scenarios is possible. The computational complexity of the attention matrix demonstrates quadratic growth concerning the number of tokens, and the acquisition of a binarized attention matrix introduces notable computational redundancy. Therefore, when dealing with a large number of tokens, replacing the attention module with specialized convolutional structures can reduce computational complexity and decrease the corresponding number of parameters, which is an effective solution to address the issue caused by an excessive number of tokens.

\section{The detailed illustration of three observations}

\begin{Proposition}
\label{lemma1}
Avoiding excessive numbers of tokens is beneficial for Binary ViT.
\end{Proposition}
% \noindent $Proof.$ For a vector $\mathbf{x}$ containing $k$ elements,$\left[ \mathrm{x}_1,\mathrm{x}_2,\cdots ,\mathrm{x}_{\mathrm{k}} \right] $, the $softmax$ operation will transfer it to the normalized vector $\widehat{\mathbf{x}}$, as shown in Eq.~\ref{softmax},
\noindent \textbf{Detailed illustration.} For a vector $\mathbf{x}$ containing $k$ elements, $\left[ x_1, x_2, \cdots, x_k \right]$, represents as one row of the attention matrix before softmax, which is the similarity vector between a token and the rest of the tokens. 

As shown in Bibert~\cite{qin2022bibert}, we assume the $\mathbf{x}$ is the $m$ row of the attention matrix before softmax, and the element of $\mathbf{x}$ can be obtained by the following,
\begin{equation}
\label{pdf}
\begin{aligned}
&x_i=\sum_{l=1}^d{B_a\left( \mathbf{Q},a_1,b_1 \right) ^{m,l}}\times B_a\left( \mathbf{K}^T,a_2,b_2 \right) ^{l,i},\\
&\mathrm{B}_a\left( \mathbf{M},a,b \right) =sign\left( \frac{\mathbf{M}-b}{a} \right) ,
\end{aligned}
\end{equation}
where $\mathrm{B}_a\left( \mathbf{M}, a,b \right)$ is the binary process of $\mathbf{M}$. $a$ and $b$ are scale factor and bias, respectively.
$l$ and $d$ are the index and number of channels of $\mathbf{Q}$ and $\mathbf{K}$, respectively.

\begin{figure}[htbp]
\centering
\includegraphics[width=3.2in]{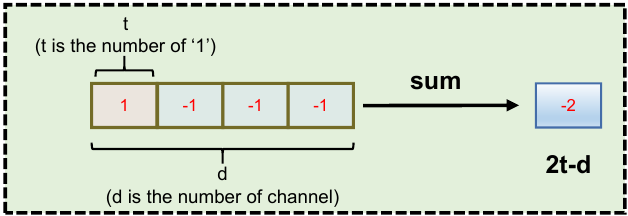}
\caption{The Schematic diagram of the process of computing $x_i$, referring to Eq.~\ref{pdf}.}
\label{figsup}
\end{figure}

\begin{figure*}[htbp]
\centering
\includegraphics[width=4.0in]{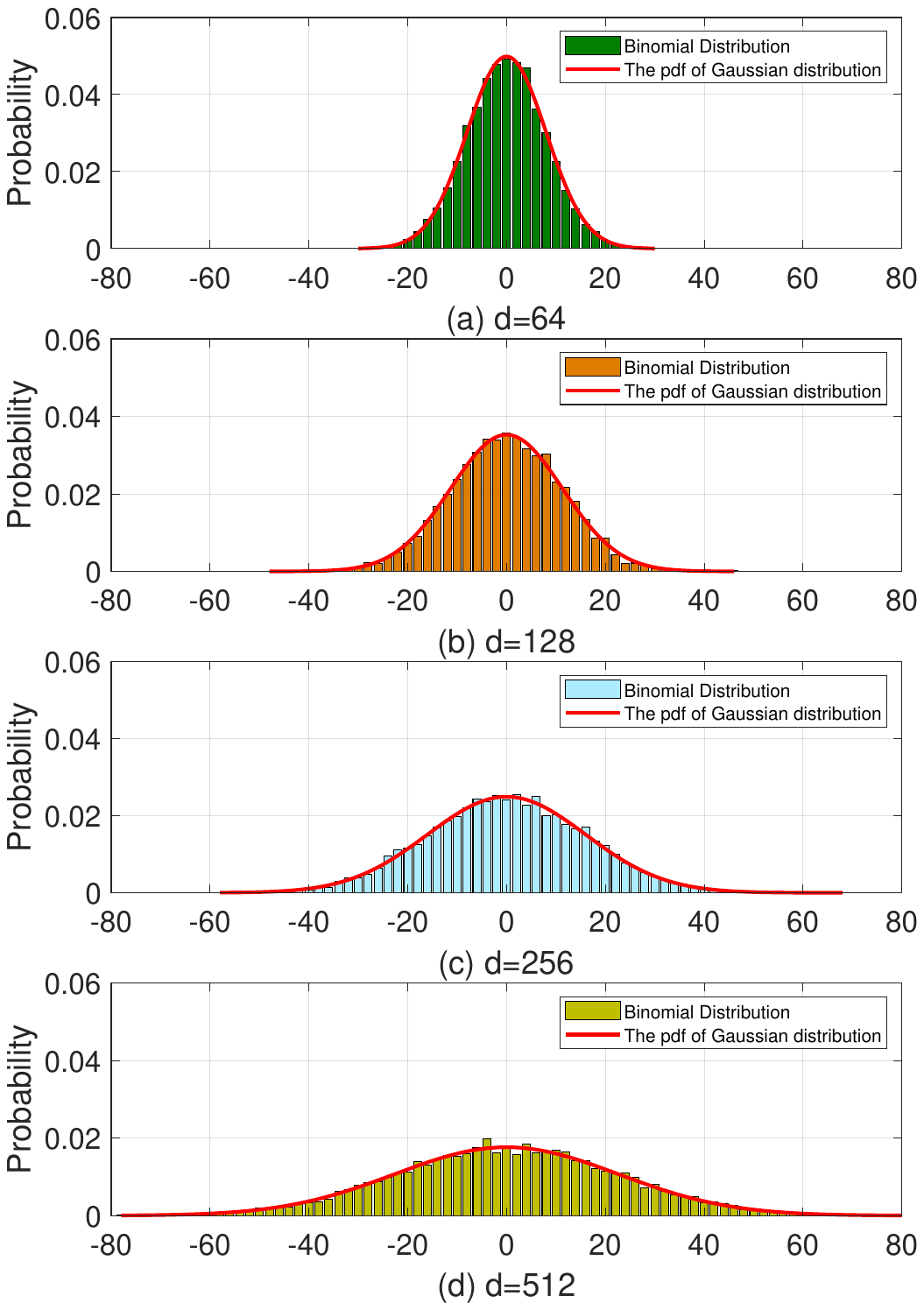}
\caption{The Schematic diagram of DeMoivre–Laplace theorem with different $d$. The histograms display the Binomial distributions with the same $p=0.5$ and different $d$. The red lines are the corresponding fitted Gaussian distributions. When $d$ increases, the Binomial distribution can be better approximated by the Gaussian distribution.}
\label{figdemo}
\end{figure*}

Let $\gamma=B_a\left( \mathbf{Q},a_1,b_1 \right) ^{m,l}\times B_a\left( \mathbf{K}^{T},a_2,b_2 \right) ^{l,i}$, thus $\gamma$ is a binary random variable taking 1 or -1, which is subject to a Bernoulli distribution with the probability of $p$ (when $\gamma =1$). Based on the binary process in Eq.~\ref{pdf}, $p$ near 0.5. Then, the probability of $x_i$, $p_{x_i}$, can be expressed as a binomial distribution. 
\begin{equation}
\label{pdf}
\begin{aligned}
p_{x_i}\left( x_i \right) =C_{d}^{t}p^{t}\left( 1-p \right) ^{d-t}
\end{aligned}
\end{equation}
where $x_i$ takes the value of $2t-d$, referring to Fig.~\ref{figsup}.   Following the DeMoivre–Laplace theorem~\cite{walker1985moivre}, $x_i$ can be well approximated by the normal distribution $\mathcal{N} \left( \mu ,\sigma ^2 \right) $ when $d$ is large enough, shown in Fig.~\ref{figdemo}. In our case, $d$ is no less than 256 (the number of channels), and the DeMoivre–Laplace theorem can be applicable very well. As the information entropy of one-dimensional Gaussian distribution is

\begin{equation}
\label{pdf2}
\begin{aligned}
&H_G(x)=-\int\limits_{-\infty}^{+\infty}{p_G\left( x \right) \ln \left( p_G\left( x \right) \right) dx}
\\
&=-\int\limits_{-\infty}^{+\infty}{\frac{1}{\sqrt{2\pi \sigma ^2}}e^{-\frac{(x-\mu )^2}{2\sigma ^2}}\cdot \ln \frac{1}{\sqrt{2\pi \sigma ^2}}e^{-\frac{(x-\mu )^2}{2\sigma ^2}}\mathrm{d}x}
\\
&=-\int\limits_{-\infty}^{+\infty}{\frac{1}{\sqrt{2\pi \sigma ^2}}e^{-\frac{(x-\mu )^2}{2\sigma ^2}}\cdot (-\ln \sqrt{2\pi \sigma ^2}-\frac{(x-\mu )^2}{2\sigma ^2})\mathrm{d}x}
\\
&=\ln \sqrt{2\pi \sigma ^2}+\int\limits_{-\infty}^{+\infty}{\frac{1}{\sqrt{2\pi \sigma ^2}}e^{-\frac{(x-\mu )^2}{2\sigma ^2}}\cdot \frac{(x-\mu )^2}{2\sigma ^2}\mathrm{d}x}
\\
&=\ln \sqrt{2\pi \sigma ^2}+\frac{1}{\sqrt{\pi}}\int\limits_{-\infty}^{+\infty}{e^{-\rho ^2}\cdot \rho ^2d\rho}
\\
&=\ln \sqrt{2\pi \sigma ^2}+\frac{1}{2}
\\
&=\frac{1}{2}\ln\mathrm{(}2\pi e\sigma ^2),
\end{aligned}
\end{equation}
where $p_G\left( x \right) =\frac{1}{\sqrt{2\pi \sigma ^2}}e^{-\frac{\left( x-\mu \right)^2}{2\sigma ^2}}$. 
\begin{figure*}[htbp]
\centering
\includegraphics[width=5.5in]{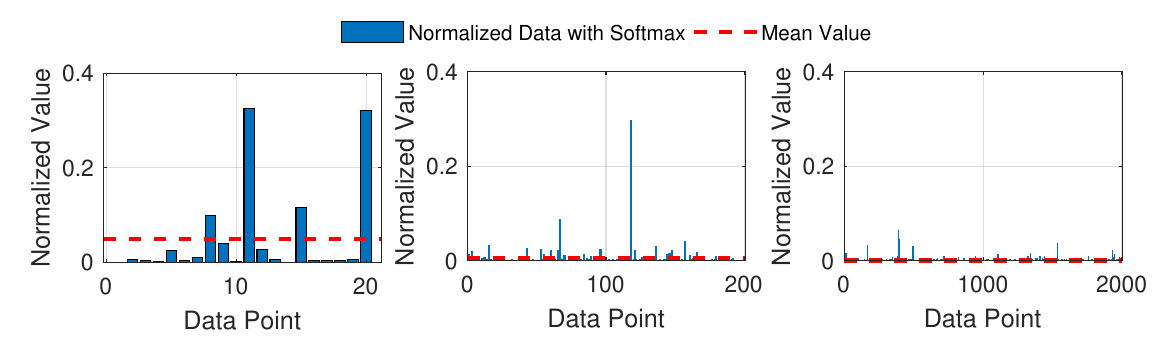}
\caption{The distribution of the sample data sets with different numbers of data. Three sets are randomly sampled from the same Gaussian distribution. (a) 20 Sample data, (b) 200 Sample data, (c) 2000 Sample data. }
\label{figts}
\end{figure*}
Assuming the elements of $\mathbf{x}$ are independent and identically distributed, the information entropy of $\mathbf{x}$ is
\begin{equation}
\label{shang0}
\begin{aligned}
H_G\left( \mathbf{x},k \right) =\frac{k}{2}\ln \left( 2\pi e\sigma^{2} \right).
\end{aligned}
\end{equation}
From Eq.~\ref{shang0}, we can find the information entropy of $\mathbf{x}$ is proportional to the token number $k$. Therefore, as the number of tokens $k$ increases, the information entropy of $\mathbf{x}$ would continuously increase.

The \textit{softmax} operation transfers vector $\mathbf{x}$ to the normalized vector $\mathbf{p}_{sof}$, as shown in Eq.~\ref{softmax},
\begin{equation}
\label{softmax}
\begin{aligned}
\mathbf{p}_{sof}^{i} =\frac{e^{x_i}}{\sum_{i=1}^k{e^{x_i}}},
\end{aligned}
\end{equation}
where $e^{x_i}$ means exponential mapping for $x_i$. $\mathbf{p}_{sof}^{i}$ represent the probabilities of the corresponding element $x_i$ in $\mathbf{x}$.

The entropy of the vector $\mathbf{x}$ after \textit{softmax} is represented by
\begin{equation}
\label{shang}
\begin{aligned}
H_s(\mathbf{x},k)&=-\sum_{i=1}^k{\mathbf{p}_{sof}^{i}}\ln(\mathbf{p}_{sof}^{i})\\
&=-\sum_{i=1}^k{\frac{e^{x_i}}{\sum_{j=1}^k{e^{x_j}}}}\ln\mathrm{(}\frac{e^{x_i}}{\sum_{j=1}^k{e^{x_j}}})\\
&=-\sum_{i=1}^k{\frac{e^{x_i}}{\sum_{j=1}^k{e^{x_j}}}}\left( x_i-\ln\mathrm{(}\sum_{j=1}^k{e^{x_j}}) \right)\\ 
&=\ln\mathrm{(}\sum_{j=1}^k{e^{x_j}})-\frac{\sum_{i=1}^k{e^{x_i}\cdot x_i}}{\sum_{j=1}^k{e^{x_j}}},\\
&=\ln\mathrm{(}k\times \frac{1}{k}\sum_{j=1}^k{e^{x_j}})-\frac{\sum_{i=1}^k{e^{x_i}\cdot x_i}}{\sum_{j=1}^k{e^{x_j}}},
\\
&=\ln\mathrm{(}k)+\ln\mathrm{(}\frac{1}{k}\sum_{j=1}^k{e^{x_j}})-\frac{\sum_{i=1}^k{e^{x_i}\cdot x_i}}{\sum_{j=1}^k{e^{x_j}}},
\end{aligned}
\end{equation}
where $\frac{\sum_{i=1}^k{e^{x_i}\cdot x_i}}{\sum_{j=1}^k{e^{x_j}}}$ is the expectation value of vector $\mathbf{x}$. $\frac{1}{k}\sum_{j=1}^k{e^{x_j}}$ is the expectation of variable $e^{x_j}$ and $e^{x_j}$ follows the log-normal distribution, a continuous probability distribution of a random variable whose logarithm is normally distributed~\cite{aitchison1957lognormal}. Therefore, we could obtain
\begin{equation}
\label{shang3}
\begin{aligned}
&H_s(\mathbf{x},k)=\ln\mathrm{(}k)+\ln\mathrm{(}\frac{1}{k}\sum_{j=1}^k{e^{x_j}})-\frac{\sum_{i=1}^k{e^{x_i}\cdot x_i}}{\sum_{j=1}^k{e^{x_j}}},\\
&= \ln\mathrm{(}k)+\ln\mathrm{(}e^{\mu +\frac{\sigma ^2}{2}})-\frac{\sum_{i=1}^k{e^{x_i}\cdot x_i}}{\sum_{j=1}^k{e^{x_j}}}, 
\end{aligned}
\end{equation}
where $x_j$ and $x_i$ follow the same Gaussian distribution $\mathcal{N} \left( \mu ,\sigma ^2 \right) $. Let $\mu_s=\frac{\sum_{i=1}^k{e^{x_i}\cdot x_i}}{\sum_{j=1}^k{e^{x_j}}}$, and we have
\begin{equation}
\label{shang4}
\begin{aligned}
H_s(\mathbf{x},k)=\ln\mathrm{(}k)+\mu +\frac{\sigma ^2}{2}-\mu_s, 
\end{aligned}
\end{equation}
The $\mu_s$ is the weighted sum of $x_i$ and the sum of weights is 1. As $k$ increases, there is an upper bound on the value of $\mu_s$.

\begin{equation}
\label{shang5}
\begin{aligned}
\mu _s=\frac{\sum_{i=1}^k{e^{x_i}\cdot x_i}}{\sum_{j=1}^k{e^{x_j}}}<\frac{\sum_{i=1}^k{e^{x_i}\cdot d}}{\sum_{j=1}^k{e^{x_j}}}=d,
\end{aligned}
\end{equation}
where $d$ is the channel number of $x_i$. Combining the Eq.~\ref{shang4} and Eq.~\ref{shang5}, the information entropy of $\mathbf{x}$ after \textit{softmax} also increases with a larger $k$. 
Meanwhile, as $k$ increases, we have
\begin{equation}
\begin{aligned}
\lim_{k\rightarrow \infty} \mathbf{p}_{sof}^{i}&=\lim_{k\rightarrow \infty} \frac{e^{x_i}}{\sum_{j=1}^{k-1}{e^{x_j}}+e^{x_i}},  \left( i\ne j \right) 
\\
&\approx \frac{e^{x_i}}{\left( k-1 \right) e^{x_i}+e^{x_i}}=\frac{1}{k},
\end{aligned}
\end{equation}
where the difference between $e^{x_i}$ and each $e^{x_j}$ can be ignored when $k\rightarrow \infty$.

Therefore, the probability distribution vector of $\mathbf{x}$ gradually approximates a uniform distribution with an increasing number of tokens. An illustrative example is shown in Fig.~\ref{figts}. Three pictures describe different numbers of samples (20, 200, and 2000) from the same Gaussian distribution, respectively. From Fig.~\ref{figts}, it is observed that the distribution is gradually approximating uniform with the number of data increasing. It is well known that a uniform distribution for the attention matrix implies that all tokens are treated equally, which undermines the effectiveness of the attention mechanism. 

From another perspective, as the number of tokens increases, the scaling factor $a$ of the binary attention matrix may become too small. This is because $\mathbf{A}_{tt}$ becomes too small when the number of tokens is large, and thus $a$ should be very small to make $\frac{\mathbf{A}_{tt}-b}{a}$ aligned to the range [0,1]. However, as shown in Eq.~\ref{eqatt}, a very small scale factor $a$ reduces the value of the final binarization result during forward propagation and results in gradient disappearance during the back-propagation process.

\begin{equation}
\label{eqatt}
\begin{aligned}
	\mathbf{Forward}&:\mathrm{B}_{att}\left( \mathbf{A}_{tt},a,b \right)\! =\\
 &a \cdot clip\left( round\left( \frac{\mathbf{A}_{tt}-b}{a} \right) ,0,1 \right),\\
	\mathbf{Backward}&:\frac{\partial L}{\partial \mathbf{A}_{tt}}=\left\{ \begin{matrix}
	a\frac{\partial L}{\partial \hat{\mathbf{A}}_{tt}}&		b\leqslant \mathbf{A}_{tt}<a +b\\
	0&		otherwise\\
\end{matrix} \right. ,
\end{aligned}
\end{equation}
where $\mathrm{B}_{att}$ represents the binary function for the full precision attention matrix $\mathbf{A}_{tt}$, and $\hat{\mathbf{A}_{tt}}$ denotes the corresponding binary attention matrix. $clip\left( x,0,1 \right)$ truncates values that fall below 0 to 0 and those above 1 to 1, effectively ensuring that the output remains within the range [0, 1]. $round$ operation maps the input to the nearest integer.

To summarize the above observation, avoiding using many tokens for the binary attention module is advisable.

%%%%%%%%%%%%%%%%%%%%%

%%%%%%%%%%%%%%%%%%%%

\begin{Proposition}
\label{lemma2}
Adding a residual connection in each binary layer is beneficial for binary ViT.
\end{Proposition}
\noindent $\textbf{Detailed illustration}.$ Layer-by-layer residual connection refers to adding a residual connection for each binarization layer in a model. The essence is that applying layer-by-layer residual connections can effectively alleviate the disappearance of activation gradients caused by the continuous superposition of gradient truncation in multiple binary layers. Meanwhile, binarization functions inherently lead to information loss in activation values, and layer-by-layer residual connections help mitigate this information loss~\cite{liu2018bi}. The current binary ViT algorithms~\cite{he2022bivit,li2024bi} only retain the residual connection outside the MLP and multi-head attention modules. Consequently, the gradient might not be fully exploited across all layers within each module of the binary ViT models. 

\begin{figure*}[htbp]
\setlength{\abovecaptionskip}{0pt}
\setlength{\belowcaptionskip}{0pt}
\centering
\includegraphics[width=5.5in]{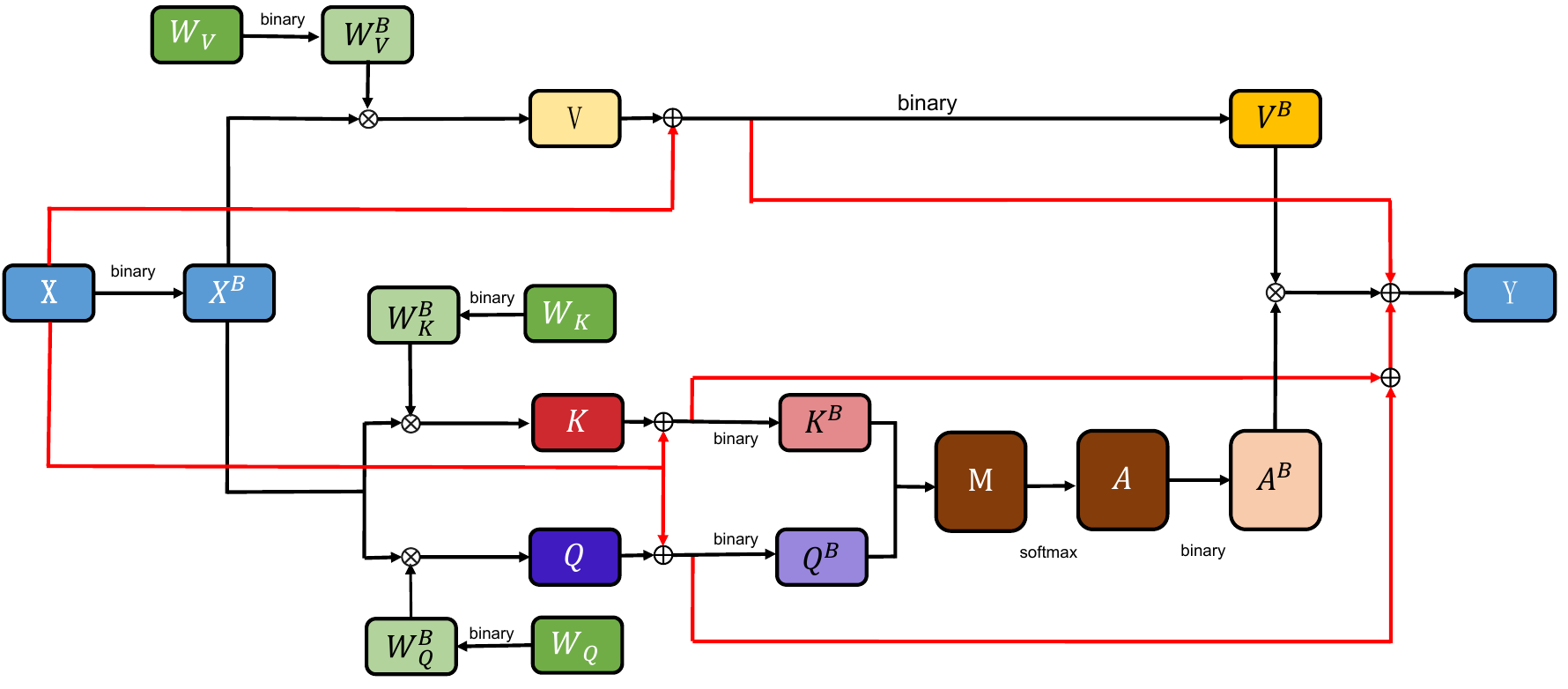}
\caption{The information flow of the binary attention of ViT. The path with the black line and arrow refers to the information flow of binary MHSA with the original architecture. The red line and arrow path denote the added residual branch, which can solve the vanishing gradient problem caused by the superposition of truncated functions without introducing too much computation. }
\label{figif}
\end{figure*}

Fig.~\ref{figif} shows one attention module. in the gradient back-propagation of current binary ViT models~\cite{he2022bivit,li2024bi}, the Jacobian of the output $Y$ is calculated with respect to the weight of the linear layer. We take the weight $W_q$ for the $Q$ tensor as an example, as shown in Eq.~\ref{eq8}, the element of $Y$ is represented by $Y^{n_l,c_i}$ and the element of $W_q$ is represented by  $W_q^{c_i,c_j}$, we have
\begin{equation}
\label{eq8}
\frac{\partial Y^{n_l,c_i}}{{\partial W_q}^{c_i,c_j}} = 
\begin{aligned}[t]
&\frac{\partial Y^{n_l,c_i}}{\partial B\left( A^{n_l,\boldsymbol{n}} \right)} \mathrel{\cdot} 
\frac{\partial B\left( A^{n_l,\boldsymbol{n}} \right)}{\partial A^{n_l,\boldsymbol{n}}} \mathrel{\cdot} 
\frac{\partial A^{n_l,\boldsymbol{n}}}{\partial M^{n_l,\boldsymbol{n}}} \mathrel{\cdot}\\
&\frac{\partial M^{n_l,\boldsymbol{n}}}{\partial B\left( Q^{n_l,c_i} \right)} \mathrel{\cdot} 
\frac{\partial B\left( Q^{n_l,c_i} \right)}{\partial Q^{n_l,c_i}} \mathrel{\cdot}\\
&\frac{\partial Q^{n_l,c_i}}{\partial B\left( W_{q}^{c_i,c_j} \right)} \mathrel{\cdot} 
\frac{\partial B\left( W_{q}^{c_i,c_j} \right)}{\partial W_{q}^{c_i,c_j}}
\end{aligned}
\end{equation}
where $\boldsymbol{n}\in \mathbb{R} ^t, l\&k\in \left[ 1,t \right] ,i\&j\in \left[ 1,d \right]$. $t$ means the token number and $d$ is the channel number. We omit each activation's batch size and head dimension for simplicity of description. $B()$ means binarization function. $M$ is the attention matrix before the \textit{softmax} operation and $\sqrt{d}$ scaling process. $A$ is the attention matrix after the \textit{softmax} process.  

Through the forward propagation path of the attention module, we can deduce the specific values of the gradients of each part of the chain rule. As shown in Eq.~\ref{eq8-1}:
%=\frac{\partial soft\max \left( M^{\boldsymbol{n}_i,\boldsymbol{n}} \right)}{\partial M^{\boldsymbol{n}_i,\boldsymbol{n}}}
\begin{equation}
\label{eq8-1}
\begin{aligned}
&\frac{\partial Y^{n_l,c_i}}{\partial B\left( A^{n_l,\boldsymbol{n}} \right)}=B\left(V^{\boldsymbol{n},c_i}\right),\\
&\frac{\partial B\left( A^{n_l,\boldsymbol{n}} \right)}{\partial A^{n_l,\boldsymbol{n}}}=\mathbf{1}_{0.5\leqslant  A^{n_l,\boldsymbol{n}} \leqslant 1},
\\
& \mathbf{1}_{0.5\leqslant A^{n_l,\boldsymbol{n}}\leqslant 1}^{n_l,n_k}=\left\{ \begin{matrix}
	1&		0.5\leqslant A^{n_l,n_k}\leqslant 1\\
	0&		others\\
\end{matrix} \right. ,
\\
&\frac{\partial A^{n_l,\boldsymbol{n}}}{\partial M^{n_l,\boldsymbol{n}}}=\frac{A^{n_l,\boldsymbol{n}}\otimes \left( 1-A^{n_l,\boldsymbol{n}} \right)}{\sqrt{d}} ,
\\
&\frac{\partial M^{n_l,\boldsymbol{n}}}{\partial B\left( Q^{n_l,c_i} \right)}=\sum_{j=1}^t{B\left( K^{c_i,n_j} \right)},
\\
&\frac{\partial B\left( Q^{n_l,c_i} \right)}{\partial Q^{n_l,c_i}}=\left\{ \begin{matrix}
	1&		\left| Q^{n_l,c_i} \right|\leqslant 1\\
	0&		others\\
\end{matrix} \right. ,
\\
&\frac{\partial Q^{n_l,c_i}}{\partial B\left( W_{q}^{c_i,c_j} \right)}=B\left(X^{n_l,c_j}\right),\\
&\frac{\partial B\left( W_{q}^{c_i,c_j} \right)}{\partial W_{q}^{c_i,c_j}}=\left\{ \begin{matrix}
	1&		\left| W_{q}^{c_i,c_j} \right|\leqslant 1\\
	0&		others\\
\end{matrix} \right. ,
\end{aligned}
\end{equation}
where $K$ is the $K$ tensor and $B\left(X\right)$ is the binary input tensor of attention module. $\otimes$ denotes Hadamard product. So, we can get the specific expression of $\frac{\partial Y^{n_l,c_i}}{{\partial W_q}^{c_i,c_j}}$ as shwon in Eq.~\ref{eq8-2}.
\begin{figure}[htbp]
\setlength{\abovecaptionskip}{0pt}
\setlength{\belowcaptionskip}{0pt}
\centering
\includegraphics[width=3.3in]{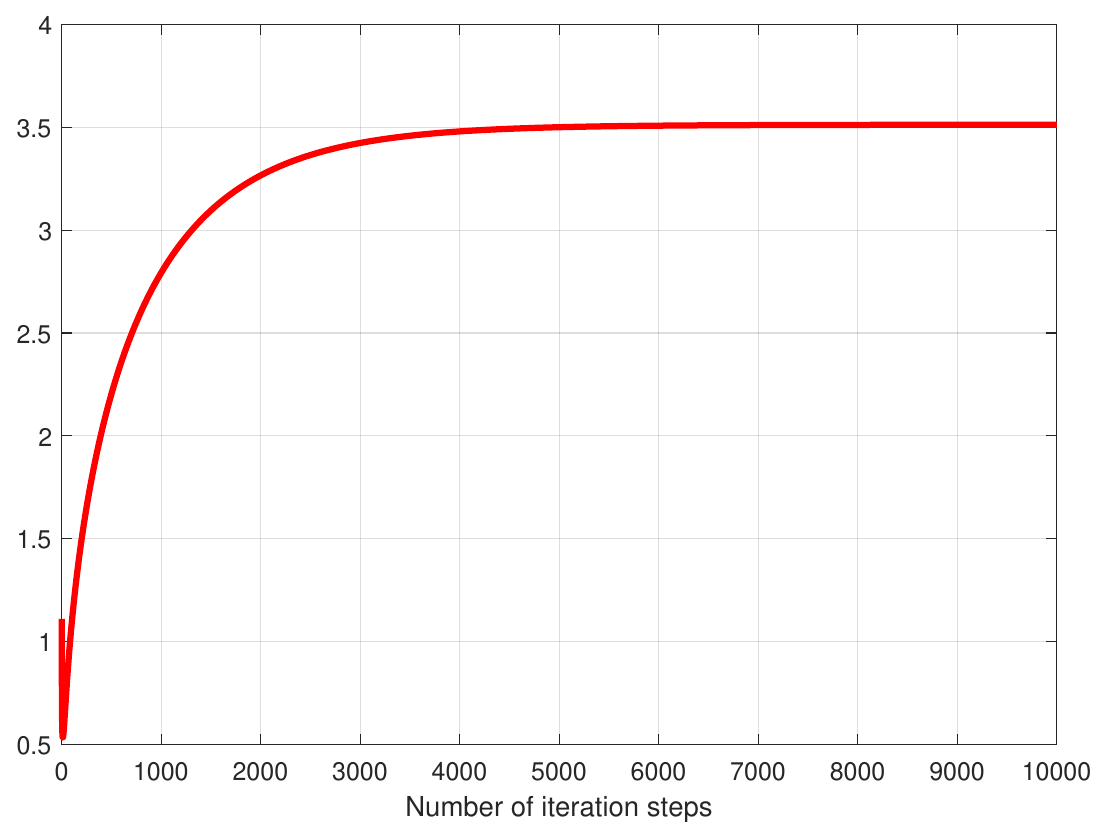}
\caption{The value of $\frac{\sqrt{\sum_{i=1}^t{\beta _{2}^{i}}}}{\sum_{i=1}^t{\beta _{1}^{i}}}$ with respect to the number of iteration steps $i$.}
\label{bztx}
\end{figure}
\begin{equation}
\label{eq8-2}
\begin{aligned}
&\frac{\partial Y^{n_l,c_i}}{{\partial W_q}^{c_i,c_j}} 
= G\!\cdot\!\frac{\partial B\left( Q^{n_l,c_i} \right)}{\partial Q^{n_l,c_i}}\!\cdot\!  B\left(X^{n_l,c_j}\right) \!\cdot\!\frac{\partial B\left( W_{q}^{c_i,c_j} \right)}{\partial W_{q}^{c_i,c_j}}, \\
&G = \sum_{k=1}^t{\left( B\left( V^{n_k,c_i} \right) ^T\cdot 1_{0.5\leqslant A^{n_l,n_k}\leqslant 1}\cdot H_k \right)},
\\
&H_k=\frac{A^{n_l,n_k}\otimes \left( 1-A^{n_l,n_k} \right)}{\sqrt{d}}\cdot B\left( K^{c_i,n_k} \right) 
\end{aligned}
\end{equation}
\normalsize
As demonstrated in Eq.~\ref{eq8-2}, the superposition of multiple binarized functions with large null range of the gradient results in vanishing gradient. To address this issue, similar to previous works~\cite{liu2018bi,le2023binaryvit}, we add a residual connection for each binary layer and attention module to avoid insufficient optimization caused by the vanishing gradients.

For example, as shown in Fig.~\ref{figif}, when we introduce a residual connection linking the $Q$ tensor and the output $Y$, the gradient of the element $Y^{n_l,c_i}$ with respect to the element $W_q^{c_i,c_j}$ is shown in Eq.~\ref{eq9}. Due to the existence of the residual link, the gradient from $Y^{n_l,c_i}$ to $Q^{n_l,c_i}$ is increased by 1, which effectively avoids the gradient disappearance problem.

\begin{equation}
\label{eq9}
\begin{aligned}
\frac{\partial Y^{n_l,c_i}}{{\partial W_q}^{c_i,c_j}}&=\left( 1+G\cdot \frac{\partial B\left( Q^{n_l,c_i} \right)}{\partial Q^{n_l,c_i}} \right) \cdot B\left(X^{n_l,c_j}\right)\\
&\cdot \frac{\partial B\left( W_{q}^{c_i,c_j} \right)}{\partial W_{q}^{c_i,c_j}},
\end{aligned}
\end{equation}
\begin{Proposition}
\label{lemma3}
The Adam optimizer enlarges the weight oscillation of binary networks in the later stages of the training process, failing to update numerous parameters effectively.
\end{Proposition}
\noindent $\textbf{Detailed illustration}.$ The previous work~\cite{liu2021adam} asserts that the regularization effect of second-order momentum in Adam is beneficial for reactivating deactivated weights,  which is more effective than SGD. However, because a significant proportion of elements in latent weights are close to zero, weight oscillation becomes a common issue in binary networks during the later stages of model training. To shed light on the underlying reason why the Adam optimizer is not well-suited for the binary network towards the end of training, we explain this phenomenon by re-examining the Adam algorithm. The computational operations involved in Adam are defined by Eq.\ref{adam} and Eq.\ref{adam2}~\cite{kingma2014adam}.
\begin{equation}
\label{adam}
\begin{aligned}
m_t&=\beta _1m_{t-1}+\left( 1-\beta _1 \right) g_t,
\\
s_t&=\beta _2s_{t-1}+\left( 1-\beta _2 \right) g_{t}^{2},
\end{aligned}
\end{equation}
where $m_t$ is first-order momentum, a weighted average of the 1st-order gradients ($g_t$) over time. $t$ is the number of iteration steps. $s_t$ is second-order momentum. $\beta _1$ and $\beta _2$ are two proportional coefficients 0.9 and 0.999, respectively. Then the first- and second-order momenta after coefficient correction, $\hat{m}_t$ and $\hat{s}_t$, and the final gradient $g_{t}^{'}$ which involved in the weight update (including learning rate $\eta$) are shown in Eq.~\ref{adam2}~\cite{kingma2014adam},
\begin{equation}
\label{adam2}
\begin{aligned}
\hat{m}_t=\frac{m_t}{1-\beta _{1}^{t}},\hat{s}_t=\frac{s_t}{1-\beta _{2}^{t}},
g_{t}^{'}=\frac{\eta \hat{m}_t}{\sqrt{\hat{s}_t}+\varepsilon},
\end{aligned}
\end{equation}
where $\varepsilon =10^{-8}$. $\beta _{1}^{t}$ and $\beta _{2}^{t}$ are the $t$ power of $\beta _{1}$ and $\beta _{2}$, respectively. Then we put Eq.~\ref{adam} into Eq.~\ref{adam2} and get the simplified form of $g_{t}^{'}$, as shown in Eq.~\ref{adam3},
\begin{equation}
\label{adam3}
\begin{aligned}
g_{t}^{'}&=\eta \frac{\frac{\sum_{i=1}^t{\left( \left( 1-\beta _1 \right) \beta _{1}^{t-i}g_i \right)}}{1-\beta _{1}^{t}}}{\sqrt{\frac{\sum_{i=1}^t{\left( \left( 1-\beta _2 \right) \beta _{2}^{t-i}g_{i}^{2} \right)}}{1-\beta _{2}^{t}}}+\varepsilon},
\end{aligned}
\end{equation}
Based on the properties of geometric sequence that $\sum_{i=1}^t{\beta ^i}=\frac{\beta \left( 1-\beta ^t \right)}{\left( 1-\beta \right)}$, Eq.~\ref{adam3} can be further simplified to Eq.~\ref{adam4}.
% \begin{equation}
% \label{adam4}
% \begin{aligned}
% &g_{t}^{'}=\eta \frac{\frac{\sum_{i=1}^t{\left( \beta _{1}^{t-i+1}g_i \right)}}{\sum_{i=1}^t{\beta _{1}^{i}}}}{\sqrt{\frac{\sum_{i=1}^t{\left( \beta _{2}^{t-i+1}g_{i}^{2} \right)}}{\sum_{i=1}^t{\beta _{2}^{i}}}}+\varepsilon}\\
% &=\eta \frac{\sqrt{\sum_{i=1}^t{\beta _{2}^{i}}}}{\sum_{i=1}^t{\beta _{1}^{i}}}\cdot \frac{\sum_{i=1}^t{\left( \beta _{1}^{t-i+1}g_i \right)}}{\sum_{i=1}^t{\left( \beta _{2}^{t-i+1}g_{i}^{2} \right)}+\sqrt{\sum_{i=1}^t{\beta _{2}^{i}}}\varepsilon}\,,
% \end{aligned}
% \end{equation}
\begin{equation}
\label{adam4}
\begin{aligned}
g_{t}^{'} &= \eta \frac{\frac{\sum_{i=1}^t{\left( \beta _{1}^{t-i+1}g_i \right)}}{\sum_{i=1}^t{\beta _{1}^{i}}}}{\sqrt{\frac{\sum_{i=1}^t{\left( \beta _{2}^{t-i+1}g_{i}^{2} \right)}}{\sum_{i=1}^t{\beta _{2}^{i}}}}+\varepsilon} \\
&= \eta \frac{\sqrt{\sum_{i=1}^t{\beta _{2}^{i}}}}{\sum_{i=1}^t{\beta _{1}^{i}}} \cdot \frac{\sum_{i=1}^t{\left( \beta _{1}^{t-i+1}g_i \right)}}{\sum_{i=1}^t{\left( \beta _{2}^{t-i+1}g_{i}^{2} \right)}+\sqrt{\sum_{i=1}^t{\beta _{2}^{i}}}\varepsilon}\,,
\end{aligned}
\end{equation}
As $t$ increases, the term $\sqrt{\sum_{i=1}^t{\beta _{2}^{i}}}\varepsilon$ gradually increase. The value of $\frac{\sqrt{\sum_{i=1}^t{\beta _{2}^{i}}}}{\sum_{i=1}^t{\beta _{1}^{i}}}$ according to the number of iteration steps is shown in Fig.~\ref{bztx}. The value of $\frac{\sqrt{\sum_{i=1}^t{\beta _{2}^{i}}}}{\sum_{i=1}^t{\beta _{1}^{i}}}$ approximately equals 3.51 when the number of iterations $i$ is larger than 5000. So the $g_{t}^{'}$ is further simplified to Eq.~\ref{adam5}. 
\begin{equation}
\label{adam5}
\begin{aligned}
g_{t}^{'}\approx 3.51\eta \times \left( \frac{\sum_{i=1}^t{\left( \beta _{1}^{t-i+1}g_i \right)}}{\sqrt{\sum_{i=1}^t{\left( \beta _{2}^{t-i+1}g_{i}^{2} \right)}}+\sqrt{\sum_{i=1}^t{\beta _{2}^{i}}}\varepsilon} \right) \,\,,
\end{aligned}
\end{equation}
where the value of $g_{t}^{'}$ is determined by the learning rate $\eta$, $\beta _1$, $\beta _2$, and the $g_{i}$. When the weight oscillation happens, the sign of gradient $g_{i}$ changes frequently, causing the numerator of Eq.~\ref{adam5} in different iteration steps to cancel each other out, while the denominator of Eq.~\ref{adam5} keeps growing (Note that the decay rate of $\beta _{2}^{t}$ is much smaller than that of $\beta _{1}^{t}$). As a result, many parameters close to 0 are deactivated in the later stages of model training. To solve this problem, we add a regularization loss function to constrain the distribution of weights to keep them away from zero.

\section{Experiment}
\subsection{Ablation study}
\paragraph{Architecture Details}~The hyper-parameters of our BHViT can be summarized in Tab.~\ref{tab1}.
\begin{table}[htbp]
    	\caption{Hyper-parameters of BHViT ($n$ is 64).}
		\label{tab1}
		\centering
  \setlength{\tabcolsep}{1pt}{
		\begin{tabular}{ccc}  %确定表格竖行格式   设置列宽度：将c换为p{ cm}
			\toprule
			Parameter     &BHViT-tiny  &BHViT-small   \\
            \midrule
			 The number of blocks & [2,2,6,2] & [3,4,8,4]\\
    
               The dimension of activation  & [n,2n,4n,8n]& [n,2n,4n,8n]\\
             
              The expand ratio of MLP & [8,8,4,4]& [8,8,4,4]\\
             
              The number of attention head  & [4,8]& [4,8]\\
			\bottomrule
		\end{tabular}}
	\end{table}
 
\begin{figure*}[htbp]
\centering
\includegraphics[width=6.5in]{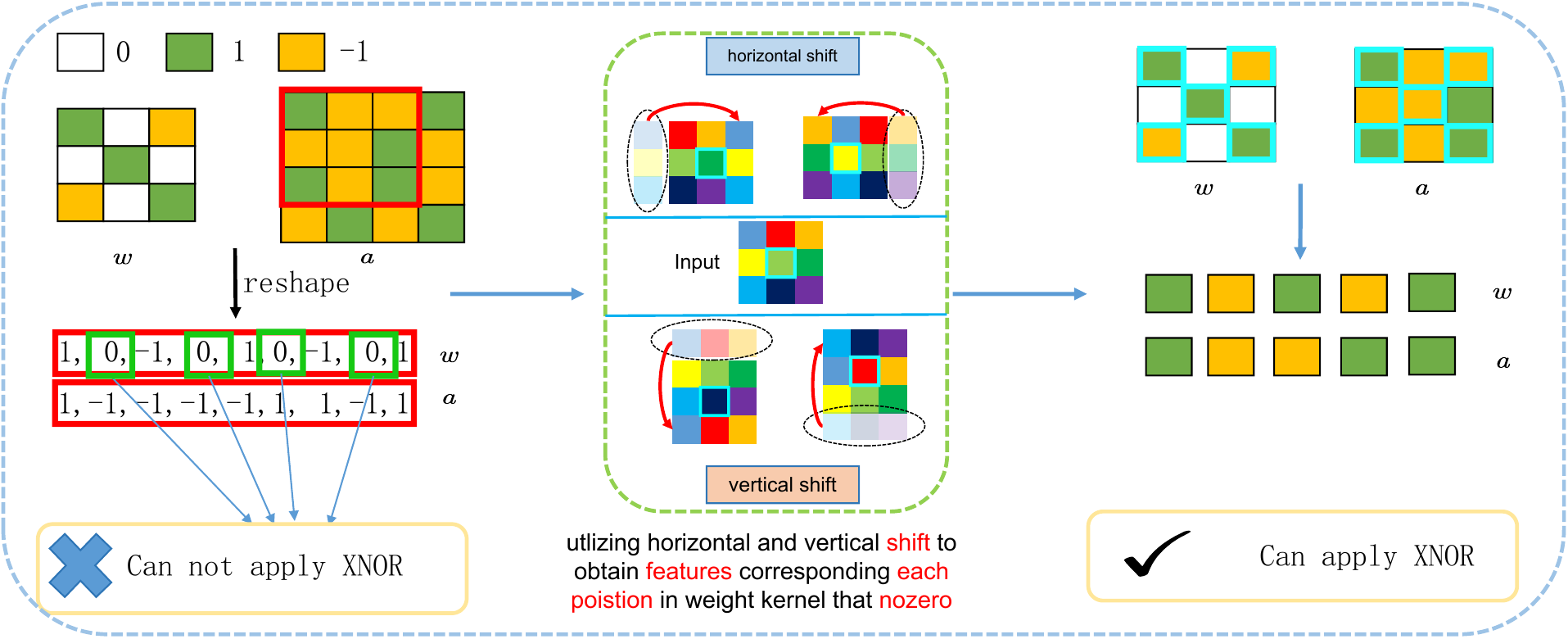}
\caption{The process of applying shift operation for binary atrous convolution layer to approximate the vector multiplication by Xnor and popcount.}
\label{figholecov}
\end{figure*} 
\paragraph{Binary atrous convolution layer} As shown in Fig.~\ref{figholecov}, due to the introduced “0” states, the binary atrous convolution layer is not suitable for deployment on binary devices. To solve this problem, we could use the shift operation proposed in section 4 to obtain the feature at the position that the corresponding weight of atrous convolution is nonzero. Then, the select feature and corresponding weight are reshaped to one dimension to apply the "xnor" and "popcount" operations instead of the multiplication between binary vectors. 

In another way, we could apply the max pooling layer (with no additional FLOPs) coordinated with standard $3\times3$ convolution to implement the convolution with different receptive fields. As shown in Tab.~\ref{tabtmv}, we conduct a performance comparison between the token mixer using dilated convolutions and the version using max pooling.\\ 
\begin{table}[htbp]
\renewcommand\arraystretch{1.3}
    	\caption{The performance of BHVIT with different version token mixer.}
		\label{tabtmv}
		\centering
         \setlength{\tabcolsep}{12pt}{
		\begin{tabular}{ccc}  %确定表格竖行格式   设置列宽度：将c换为p{ cm}
			\Xhline{1pt}
			  Network & Token mixer & Top1(\%)  \\
			\Xhline{1pt}	
       BHViT & Dilated Convolution & 70.1 \\
       BHViT & Max pooling & 69.8 \\
		\Xhline{1pt}
		\end{tabular}}
	\end{table}
 
According to Tab.~\ref{tabtmv}, each version of the token mixer has its advantages. Dilated convolutions obtain higher classification accuracy, but deploying this setting requires pre-processing for the activation. Deploying the token mixer with max pooling is relatively easy, but obtaining a relatively lower accuracy.

\paragraph{The ablation study about the latency}
To obtain a latency result comparison between the full precision BHViT and the corresponding binary version, we first transfer the Pytorch code of BHViT to the ONNX version. Then, we utilize the BOLT toolbox~\cite{bolt} to implement our method to the edge device based on an ARM Cortex-A76 CPU (without cuda). The result is shown in the Tab.~\ref{tablatency}. Due to the lack of optimization and deployment methods for the specific modules in the ViT structure, the acceleration results of BHViT cannot achieve an ideal acceleration state the same as the BNN. Therefore, further deployment techniques must be developed to show the full advantages of binary vision transformers on edge devices.
\begin{table}[htbp]
\renewcommand\arraystretch{1.5}
    	\caption{The latency result of the full precision BHViT and binary BHViT.}
		\label{tablatency}
		\centering
         \setlength{\tabcolsep}{18pt}{
		\begin{tabular}{ccc}  %确定表格竖行格式   设置列宽度：将c换为p{ cm}
			\Xhline{1pt}
			  Network & W/A (bit) & Latency (ms)  \\
			\Xhline{1pt}
            Si-BiViT & 32/32 & 1029 \\
       BHViT & 32/32 & 612 \\
       Si-BiViT & 1/1 & 863 \\
       BHViT & 1/1 & 157 \\
		\Xhline{1pt}
		\end{tabular}}
	\end{table}

\paragraph{The impact of different architecture:} In this subsection, we compare the performance differences of three variants of ViT architectures before and after the binarization process. As shown in Tab.~\ref{tabdiff}, The accuracy differences of the three network structures before and after binarization are 30.4\%, 12.2\%, and 10.9\%, respectively. Compared with DeiT-Small and BinaryViT, the network architecture of the proposed BHViT is more suitable for binarization. 
\begin{table}[htbp]
\renewcommand\arraystretch{1.0}
    	\caption{The performance difference of three variants of ViT.}
		\label{tabdiff}
		\centering
         \setlength{\tabcolsep}{7pt}{
		\begin{tabular}{ccccc}  %确定表格竖行格式   设置列宽度：将c换为p{ cm}
			\Xhline{1pt}
			  Network& Binary method & W/A  & Top1(\%)  \\
			\Xhline{1pt}	
			\multirow{2}*{DeiT-Small~\cite{touvron2021training}}& \multirow{2}*{ReActNet~\cite{liu2020reactnet}}   &32-32&79.9 \\
            \cdashline{3-4}
             & &1-1&49.5\\
              \hline
              \multirow{2}*{BinaryViT~\cite{le2023binaryvit}}& \multirow{2}*{BinaryViT~\cite{le2023binaryvit}}  &32-32&79.9 \\
              \cdashline{3-4}
              &&1-1&67.7\\
              \hline
			\multirow{2}*{Ours}&   \multirow{2}*{Ours}   & 32-32&79.3 \\
            \cdashline{3-4}
             & &1-1&68.4\\
		\Xhline{1pt}
		\end{tabular}}
	\end{table}

\end{document}